\documentclass[journal]{IEEEtran}
\usepackage{graphicx} 
\usepackage{subfigure}
\usepackage[ruled]{algorithm2e}
\usepackage{multirow,subfigure,pifont,hyperref}
\usepackage{verbatim}
\RequirePackage{amsmath}[2016/11/05]

\ifCLASSINFOpdf
\else
\fi
\begin{document}
%
\title{TCDesc: Learning Topology Consistent Descriptors for Image Matching}
%
%
%

\author{Honghu Pan,	Fanyang Meng, Nana Fan, Zhenyu He*,~\IEEEmembership{Senior Member,~IEEE,}
\thanks{*Corresponding author.}
\thanks{E-mail of Corresponding author: zhenyuhe@hit.edu.cn.}
\thanks{H. Pan, N. Fan and Z. He are with School of Computer Science and Technology, Harbin Institute of Technology, Shenzhen, China. F. Meng is with Pengcheng Lab, Shenzhen, China.}
}

\maketitle

\begin{abstract}
The constraint of neighborhood consistency or local consistency is widely used for robust image matching. In this paper, we focus on learning neighborhood topology consistent descriptors (TCDesc), while former works of learning descriptors, such as HardNet and DSM, only consider point-to-point Euclidean distance among descriptors and totally neglect neighborhood information of descriptors.
To learn topology consistent descriptors, first we propose the linear combination weights to depict the topological relationship between center descriptor and its $\boldsymbol{k}$NN descriptors, where the difference between center descriptor and the linear combination of its $\boldsymbol{k}$NN descriptors is minimized.
Then we propose the global mapping function which maps the local linear combination weights to the global topology vector and define the topology distance of matching descriptors as $ \boldsymbol{l1}$ distance between their topology vectors.
Last we employ adaptive weighting strategy to jointly minimize topology distance and Euclidean distance, which automatically adjust the weight or attention of two distances in triplet loss.
Our method has the following two advantages:
(1) We are the first to consider neighborhood information of descriptors, while former works mainly focus on neighborhood consistency of feature points;
(2) Our method can be applied in any former work of learning descriptors by triplet loss.
Experimental results verify the generalization of our method:
We can improve the performances of both HardNet and DSM on several benchmarks.
\end{abstract}


%
\IEEEpeerreviewmaketitle

\section{Introduction}
\IEEEPARstart{I}{mage} matching~\cite{SIFT,GMS,jiang2019feature} is a fundamental computer vision problem and the crucial step in augmented reality (AR)~\cite{ARsurvey,markerlessAR} and simultaneous localization and mapping (SLAM)~\cite{ORB-SLAM,ORB-SLAM2}, which usually consists of two steps: detecting the feature points and matching feature descriptors. The robust and discriminative descriptors are essential for accurate image matching.
Early works mainly focus on the handcrafted descriptors.
SIFT~\cite{SIFT} maybe is the most successful handcrafted descriptor and has been proven effective in various areas~\cite{csurka2004visual,zheng2017sift,wide_baseline}, which is scale-invariant and orientation-invariant benefited from the Difference-of-Gaussian (DoG) scale space and the assigned main orientation respectively.
Meanwhile, the binary descriptors~\cite{Brief} are proposed to reduce storage and accelerate matching, where Hamming distance is employed to compare two binary descriptors.
However, handcrafted descriptors are not robust enough since they only consider pixel-level information and lack high-level semantic information. 

Recently with the successful application of convolutional neural networks (CNNs) in multiple fields~\cite{lecun2015deep,girshick2015fast,bertinetto2016fully}, researchers~\cite{MatchNet,DeepDesc,DeepCompare,L2Net,HardNet} try to learn descriptors directly from image patch by using CNNs.
Specifically, CNNs take image patches cropped around feature points as input and take the representation vector of last layer as the learned descriptors.
Recent works~\cite{HardNet,ExpLoss,DSM} mainly focus on learning descriptors using triplet loss~\cite{FaceNet} to encourage Euclidean distance of negative samples is a margin larger than that of positive samples, where negative samples and positive samples denote the non-matching descriptors and matching descriptors respectively.
During CNNs' training, Euclidean distance of matching descriptors is minimized and that of non-matching descriptors is maximized.

\begin{figure}[!t]
\centering
\subfigure[HardNet~\cite{HardNet} or DSM~\cite{DSM}]{\includegraphics[width=0.23\textwidth]{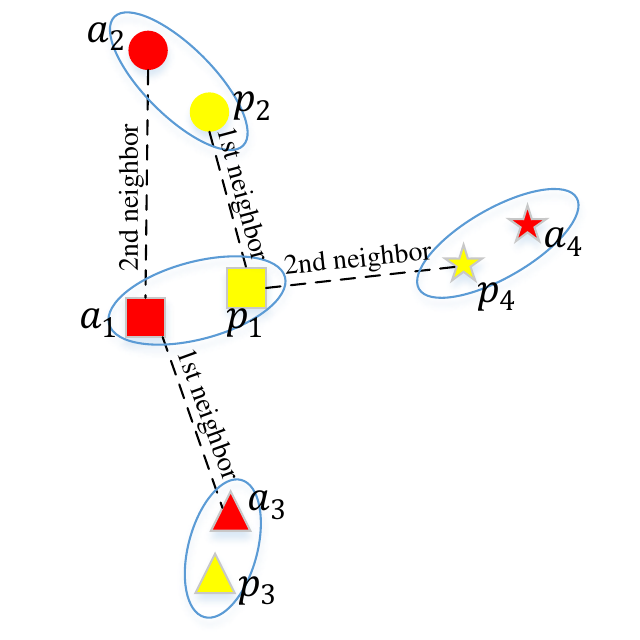}
\label{distribution-comparison-a}}
\subfigure[our method]{\includegraphics[width=0.23\textwidth]{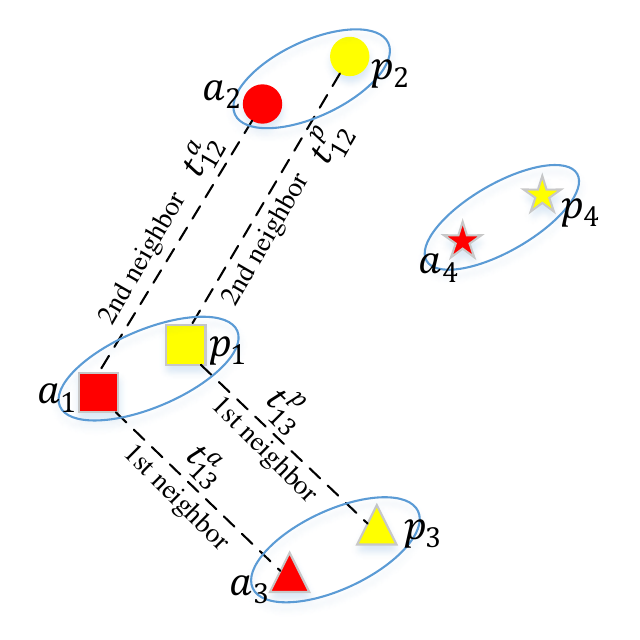}
\label{distribution-comparison-b}}
\caption{
	Distribution of descriptors learned by (a) former works of learning descriptors by triplet loss, such as HardNet~\cite{HardNet} and DSM~\cite{DSM}, and (b) our method.
	Marks with the same shape denote matching descriptors (e.g. $a_1$ and $p_1$) and marks with the same color denote descriptors from the same set (e.g. $a_1$, $a_2$, $a_3$, and $a_4$).
	In (a), neighborhood topology of descriptor $a_1$ is quite different with that of $p_1$, which results from the neglect of neighborhood information of former works~\cite{HardNet,ExpLoss,DSM}.
	In (b), our method can learn neighborhood topology consistent descriptors by minimizing the neighborhood topological difference of matching descriptors.
	}
\label{distribution-comparison}
\end{figure}

However, as shown in Fig.~\ref{distribution-comparison-a}, triplet loss of former works only considers Euclidean distance between descriptors and completely neglects the neighborhood information of descriptors, which results in the topology difference between matching descriptors.
Neighborhood consistency is wide adopted by former works~\cite{meng2015feature,GMS} for more robust image matching, which assume the local neighborhood structures of two matching feature points should be as similar as possible.
Motivated by above idea, we try to learn the neighborhood topology consistent descriptors as Fig.~\ref{distribution-comparison-b} by imposing the penalty to the topological difference of matching descriptors.

To this end, we first propose some assumptions about the distribution of descriptors in two learned descriptor sets. Specifically, for matching descriptors $a_i$ and $p_i$, we assume (1)$k$NN descriptors of $a_i$ match that of $p_i$, (2)topological relationship between $a_i$ and its $k$NN descriptors should be similar with that between $p_i$ and $k$NN descriptors of $p_i$.
Former works~\cite{LE,li2019neighborhood} usually employ the hard weight and heat kernel similarity to indicate the topological relationship between two samples, however, we figure out that they both have their own disadvantages: The hard weight is not differential and the heat kernel similarity consists of tunable hyper-parameter.
We then propose the linear combination weights to measure the topological relationship between the center descriptor and its $k$-nearest neighbor ($k$NN) descriptors.
Specifically, the difference between center descriptor $a_i$ or $p_i$ and the linear combination of its $k$NN descriptors is minimized, where the linear combination weights have the closed-form solutions because above optimization question is the Least Squares problem.

The linear combination weights are defined in a small local region, then we propose the global mapping function which maps the linear combination weights to the global topology vector.
The length of topology vector is equal to $n$, the training batch size, and it is the sparse vector with only $k$ non-zero elements.
In this paper, topology distance between two matching descriptors is defined as the $l1$ distance of their topology vectors, then we can learn topology consistent descriptors by minimizing topology distance of matching descriptors.

To learn more robust and discriminative descriptors, we jointly minimize the topology distance and Euclidean distance of matching descriptors  by modifying the distance of positive sample in triplet loss to the weighted sum of topology distance and Euclidean distance.
Otherwise, we propose the adaptive weighting strategy to adjust their respective weight: the more stable neighborhood set of descriptor is, the lager weight we assign to the topology distance, where the stability of neighborhood set is proportional to the matching pairs within two neighborhood sets.
Our method modifies and consummates the distance measure of positive samples for triplet loss, which means our method can be applied in any other algorithms~\cite{HardNet,DSM,ExpLoss} of learning descriptors by triplet loss.
The generalization of our method is verified in several benchmarks in Section~\ref{Extensive_Experiments}.

The contributions of this paper are four-fold:
\begin{itemize}
	\item We are the first to consider neighborhood information of learning-based descriptors, where the neighborhood topological relationship between center descriptor and its $k$NN descriptors is depicted by our linear combination weights;
	\item We propose the global mapping function to map the local linear combination weights to the global topology vector, and then define the topology distance of matching descriptors as $l1$ distance between their topology vector;
	\item We propose the adaptive weighting strategy to jointly minimize topology distance and Euclidean distance, which automatically adjust the weight or attention of two distances in triplet loss;
	\item The experimental results verify the generalization of our method. We test our method on the basis of HardNet~\cite{HardNet} and DSM~\cite{DSM}, and experimental results show our method can improve their performances in several benchmarks.
\end{itemize}

The rest of the paper is organized as follows. 
Section~\ref{RW} reviews some related works about the learning-based descriptors and works considering neighborhood information in unsupervised or supervised learning and image matching.
Section~\ref{method} presents our proposed method, including our novel linear combination weights, the topology distance and the adaptive weighting strategy.
Section~\ref{experiments} shows the experimental results, including ablation experiments and extensive experiments in several benchmarks.
Last Section~\ref{conclusion} draws the brief conclusions.

\section{Related Works}
\label{RW}
In this section,
we first briefly introduce several algorithms on learning descriptors in Section~\ref{LbD},
and then illustrate the effectiveness and wide application of neighborhood consistency in Section~\ref{NC}.

\subsection{Learning-based Descriptors}
\label{LbD}
Perhaps SIFT~\cite{SIFT} is the most successful and widely used handcrafted descriptor, however, all handcrafted descriptors, including SIFT~\cite{SIFT}, LIOP~\cite{LIOP}, GLOHP~\cite{GLOH}, DAISYP~\cite{DAISY}, DSP-SIFTP~\cite{DSP-SIFTP} and BRIEF~\cite{Brief} are not robust enough as they only consider the pixel-level information and neglect the high-level semantic information.
With the successful application of deep learning on various fields~\cite{tian2020coarse,liu2019learning}, researchers try to learn descriptors using CNNs directly from the image patch around feature points.
DeepCompare~\cite{DeepCompare} and MatchNet~\cite{MatchNet} learn the pairwise matching probabilities directly from image patches instead of the discriminant descriptors, in which MatchNet uses Siamese CNN with three convolutional layers and DeepCompare explores several CNN architectures.
L2-Net~\cite{L2Net} proposes a deeper CNN with seven convolutional layers to extract the semantic information and a Local Response Normalization layer to normalize descriptors, and this architecture is employed by many works~\cite{HardNet,DSM,SOSNet} including ours due to its effectiveness.

HardNet~\cite{HardNet} first introduces triplet loss~\cite{FaceNet} to learn descriptors which encourages Euclidean distance of nearest non-matching descriptors is a margin lager than that of matching pairs.
CDbin~\cite{CDbin} combines triplet loss and other three losses to learn more robust descriptors and explores the performance of descriptors with different lengths.
Followed by focal loss~\cite{focalloss}, Exp-TLoss~\cite{ExpLoss} proposes the exponential triplet loss to focus more on hard positive samples to accelerate CNNs' training.
SOSNet~\cite{SOSNet} proposes the second-order regularization to learn more robust descriptors by minimizing the edge similarity between matching descriptors.
DSM~\cite{DSM} replaces the hard margin in triplet loss by Cumulative Distribution Function-based soft margin which uses more negative samples to update CNNs.

However, above methods which use triplet loss to learn descriptors all focus on the point-to-point Euclidean distance measure and totally neglect the neighborhood information of descriptors.
In this paper, we can learn neighborhood consistent descriptors by minimizing topology difference of matching descriptors.

\begin{figure*}[ht]
	\centering
	\subfigure[SIFT~\cite{SIFT}]{\includegraphics[width=0.32\textwidth]{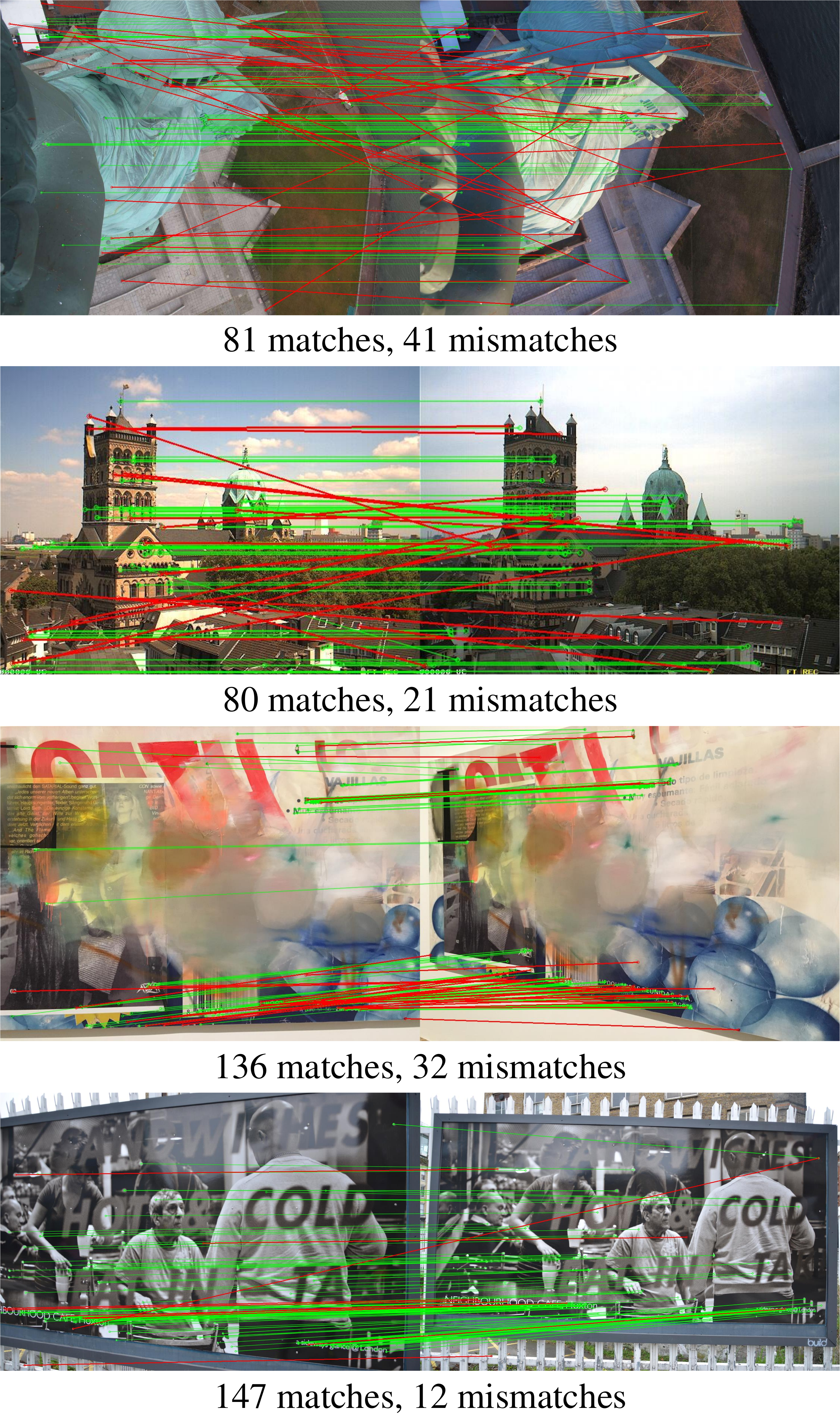}}
	\subfigure[BRISK~\cite{BRISK}+DSM~\cite{DSM}]{\includegraphics[width=0.32\textwidth]{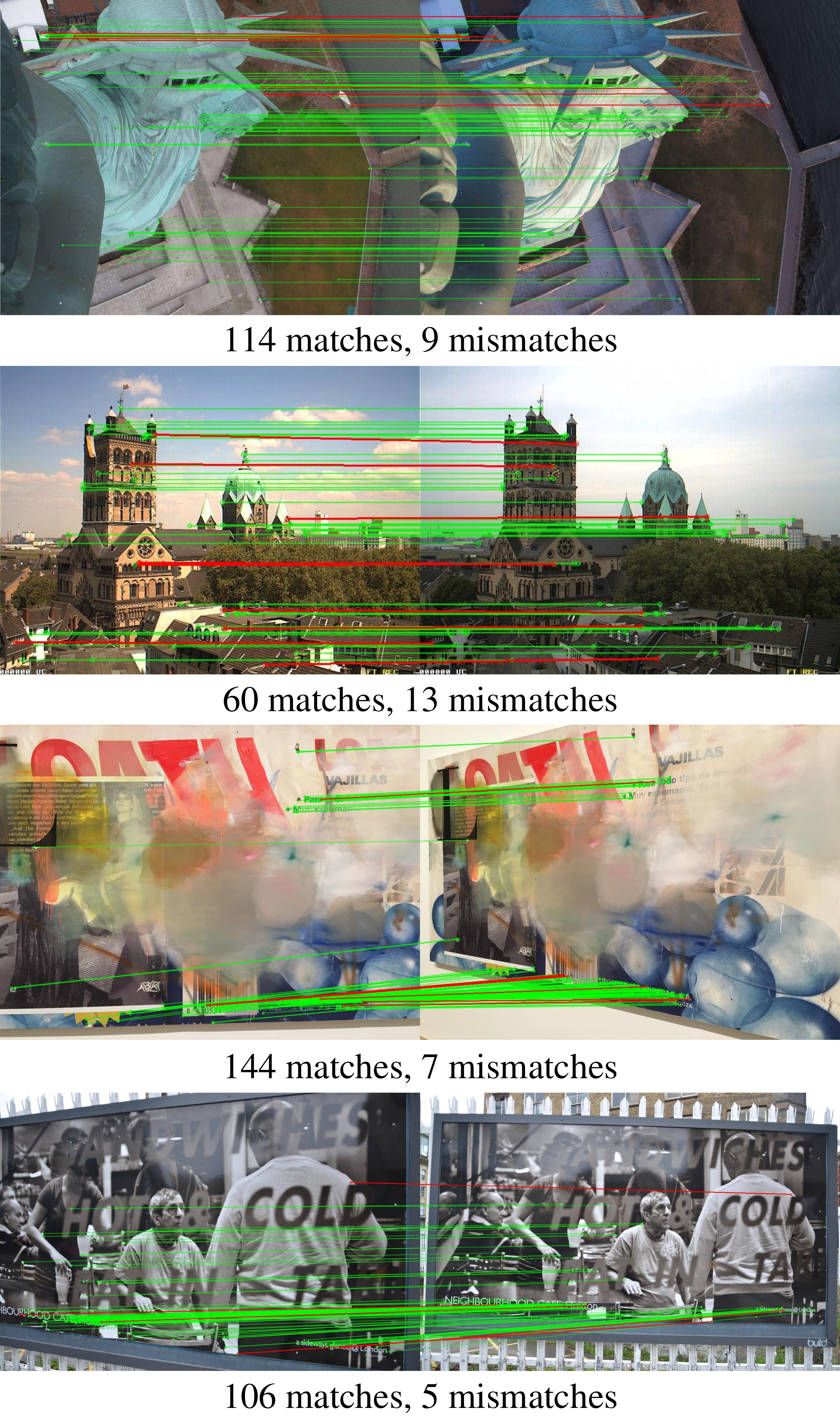}}
	\subfigure[BRISK~\cite{BRISK}+ours]{\includegraphics[width=0.32\textwidth]{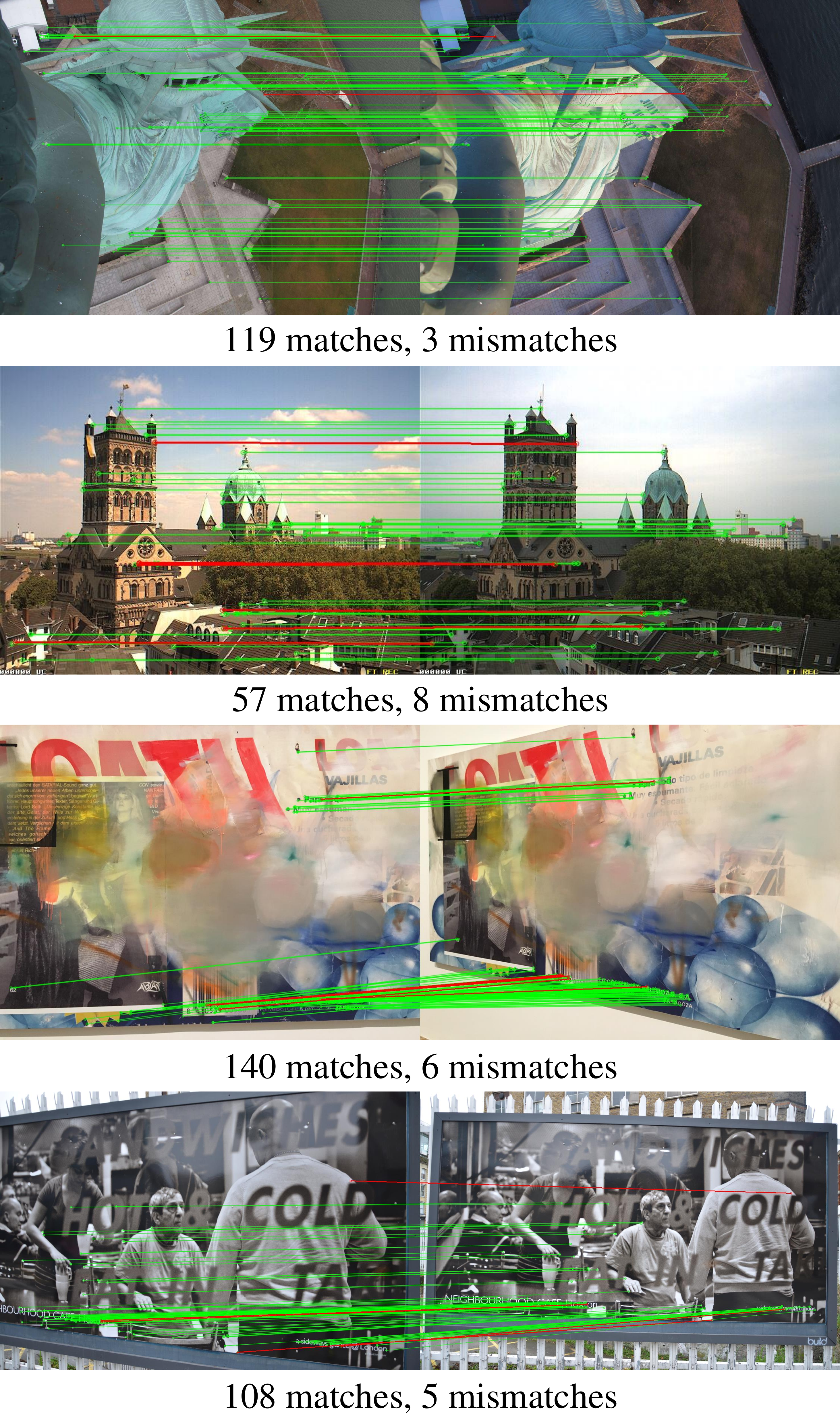}}
	\caption{Some examples of image matching on HPatches benchmark~\cite{HPatches}, where the left column, middle column and right column present the matching results by SIFT~\cite{SIFT}, DSM~\cite{DSM} and our TCDesc respectively. The green lines denote the right matches and the red lines denote wrong matches or mismatches. The nuisance of top two rows is illumination and that of below two rows is viewpoint. BRISK~\cite{BRISK} is employed to detect feature points for DSM~\cite{DSM} and our TCDesc. Our TCDesc performs similarly with DSM~\cite{DSM} under  the nuisance of viewpoint and outperforms it under the nuisance of illumination.}
	\label{fig_matches}
\end{figure*}

\subsection{Neighborhood Consistency}
\label{NC}
Neighborhood consistency or local consistency is widely applied in both unsupervised learning and supervised learning.
As the classical unsupervised and non-linear data dimensionality reduction methods, manifold learning~\cite{ISOmap,LLE,LE,LPP} tries to preserve the local relative relationship of high-dimensional data in the low-dimensional data.
In Locally Linear Embedding (LLE)~\cite{LLE}, the local relative relationship indicates neighborhood reconstruction coefficients, which are calculated by minimizing the neighborhood reconstruction error.
In Laplacian Eigenmaps (LE)~\cite{LE}, the local relative relationship indicates the hard weights or heat kernel similarity between the center sample and its neighborhood samples.
Recently, Sabokrou~\cite{karami2017image} proposes the Neighborhood-Relational Encoding for unsupervised representation learning, which determines the loss of Encoder-Decoder structure by the neighborhood relational information.
Similarly, Li~\cite{li2019neighborhood} employs the neighborhood information as the constraint to learn hash representation for large scale image retrieval.

Besides the effective constraints of local or neighborhood consistency in unsupervised learning, neighborhood consistency also shows its privilege in supervised or semi-supervised learning.
Belkin~\cite{belkin2006manifold} proposes an universal semi-supervised framework with the manifold consistency regularization, which is proven effective in Least Squares and Support Vector Machine (SVM).
NPNN~\cite{NPNN} proposes a non-linear method for data-driven fault detection, which considers the local geometrical structure of training data in neural networks.
PointWeb~\cite{PointWeb} learns 3D point cloud representation with integrating neighborhood information of each points.

Neighborhood consistency is proven effective in image matching as well.
Meng~\cite{meng2015feature} proposes the spatial order constraints bilateral-neighbor vote (SOCBV) to remove outliers with considering the $k$NN feature points of matching pairs.
GMS~\cite{GMS} depicts the neighborhood matching by statistical likelihood of the matching number in a region to enable ultra-robust matching.
LPM~\cite{LPM} attempts to remove mismatches with local neighborhood structures of potential true matches maintained.
Above works mainly focus on the neighborhood consistency of feature points, while we constrain similar neighborhood topology of matching descriptors.
As shown in Fig.~\ref{fig_matches}, compared with DSM~\cite{DSM}, our TCDesc is more robust to the illumination change with less mismatches.

\section{Methodology}
\label{method}
In this section, we first propose some assumptions about the distribution of descriptors in Section~\ref{BNC}, which serve as some basic neighborhood constraints to guide our model.
Then we present our linear combination weights in Section~\ref{TW}.
In Section~\ref{TM} we define the topology vector and topology distance between matching descriptors.
Last we propose the adaptive weighting strategy to fuse topology distance and Euclidean distance in Section~\ref{AWS}. 

In order to illustrate our model and method more clearly, here we present implications or definitions of some notations:
\begin{itemize}
	\item $a_i$, $p_j$ -- The learned descriptors, where descriptors with the same subscript are matching descriptors, such as $a_i$ and $p_i$;
	\item $t_{ij}^a$, $t_{ij}^p$ -- The topology weights which depict topological relationship between $a_i$ and $a_j$ and that between $p_i$ and $p_j$ respectively.
	\item $a_{ij}$, $p_{ij}$ -- The neighborhood descriptors of $a_i$ and $p_i$ respectively.
	\item $N(a_i)$, $N(p_i)$ -- The sets of neighborhood descriptors of $a_i$ and $p_i$, which consist of $k$NN descriptors of $a_i$ and $p_i$ respectively.
\end{itemize}

\subsection{Basic Neighborhood Constraints}
\label{BNC}
Recent works~\cite{HardNet,DSM,SOSNet} adopt L2-Net~\cite{L2Net} to learn discriminative descriptors directly from image patches. Specifically, a batch of training data generates the corresponding descriptors $\chi=\{A;P\}$, where $A=\{a_1, a_2,...,a_n\},P=\{p_1, p_2,...,p_n\}$ and $n$ is the batch size.
Normally descriptor vectors are unit-length and 128-dimensional as SIFT~\cite{SIFT}.
Note that descriptors from two sets with the same subscripts are a matching pair, such as $a_i$ and $p_i$, while descriptors with the different subscripts form the non-matching pairs. During CNNs' training, Euclidean distance of matching descriptors is minimized and that of non-matching descriptors is maximized.

Ideally, Euclidean distance of matching descriptors is equal to $0$, i.e. $\| a_i - p_i \|_2 = 0$, and Euclidean distance of non-matching descriptors should be as large as possible.
Under this circumstance, distribution of descriptors $a_i$ in set $A$ is the same as descriptors $p_i$ in set $P$.
On the basis of above conclusion, two assumptions are proposed about the neighborhood set of descriptors:
\\[5pt]
\noindent\textbf{Assumption 1 (Neighborhood Matching):} \textit{For $a_i$ in $A$ and $p_i$ in $P$, neighborhood descriptors of $a_i$ match neighborhood descriptors of $p_i$.}
\\[5pt]
\noindent Specifically, note $a_i$ and $p_i$, $a_j$ and $p_j$ are two matching pairs, then $p_j$ should be in the neighborhood set of $p_i$ if $a_j$ is in the neighborhood set of $a_i$. Similarly, $p_j$ should be far away from $p_i$ if $a_j$ is not in the neighborhood set of $a_i$.
\\[5pt]
\noindent\textbf{Assumption 2 (Neighborhood Consistency):} \textit{For $a_i$ in $A$ and $p_i$ in $P$, the topological relationship between $a_i$ and its neighborhood descriptors should be similar with that between $p_i$ and its neighborhood descriptors.}
\\[5pt]
\noindent We note that \textbf{Assumption 1} is the pre-condition of \textbf{Assumption 2} and \textbf{Assumption 2} is the further conclusion of \textbf{Assumption 1}.
Assume $a_j$ is in the neighborhood set of $a_i$ and $p_j$ is in the neighborhood set of $p_i$, then \textbf{Assumption 2} requires topological relationship between $a_i$ and $a_j$ should be equal to that between $p_i$ and $p_j$.
In this paper, we employ the topology weight $t_{ij}^a$ or $t_{ij}^p$ to measure the topological relationship between two descriptors.
According to \textbf{Assumption 2}, one of our goals is to minimize the gap of two topology weights:
\begin{equation}
<a_i,a_j,p_i,p_j> = \mathop{\arg\min}_{a_i,a_j,p_i,p_j} \left| t_{ij}^a-t_{ij}^p \right|
\label{eq_mins}
\end{equation}
Where $t_{ij}^a$ and $t_{ij}^p$ are the topology weights.

\subsection{Neighborhood Topology Weights}
\label{TW}
In Eq.~\ref{eq_mins}, the topology weights $t_{ij}^a$ and $t_{ij}^p$ depict the topological relationship between descriptor $a_i$ and $a_j$ and that between $p_i$ and $p_j$. 
In this section, we first review the former topology weights and point out their disadvantages, then we present our novel linear combination weights.

In former works, hard weight and heat kernel similarity are commonly used to measure the topology relationship between two samples:

\noindent\textbf{Hard Weight:} Hard weight~\cite{LE} is also called the binary weight.
For two descriptors $a_i$ and $a_j$, the hard weight $h(a_i,a_j)$ is equal to $1$ if $a_j$ is in the neighborhood set of $a_i$ and equal to $0$ if not:
\begin{equation}
h(a_i,a_j)=\left\{
\begin{aligned}
&1, \ \ a_j \in N(a_i)\\
&0, \ \ otherwise\\
\end{aligned}
\label{eq_binary}
\right.
\end{equation}
The definition of hard weight is very simple without consuming large amount of computation.
However, the hard weight only depicts whether $a_j$ is in the neighborhood set of $a_i$ and ignores the relative position relationship between  $a_j$ and $a_i$.
Otherwise, Eq.~\ref{eq_binary} is a \textit{discrete function}, whose gradient is undefined at the boundary of neighborhood set and equal to $0$ everywhere else.
So a differentiable proxy is required for training purpose if we choose hard weight or binary weight as our topological weight.

\noindent\textbf{Heat Kernel Similarity:} The heat kernel similarity $s$ is widely adopted by former works~\cite{LE,GMC}, which is defined as the exponential value of minus distance between center descriptor $a_i$ and its $k$NN descriptor $a_j$:
\begin{equation}
s(a_i,a_j)=\left\{
\begin{aligned}
&exp\{- \frac{\| a_i - a_j \|_2}{t} \}, \ \ a_j \in N(a_i)\\
&0, \ \ \ \ \ \ \ \ \ \ \ \ \ \ \ \ \ \ \ \ \ \ \ \ otherwise\\
\end{aligned}
\label{eq_hks}
\right.
\end{equation}
In above equation, the smaller distance contributes the lager similarity or topology weight.
However, the heat kernel similarity has the following two disadvantages:
Firstly, the hyper-parameter $t$ is hard to determined, for example, $s(a_i,a_j)$ would be close to $1$ if $t$ is a very large number and close to $0$ if $t$ is very small.
Secondly, the heat kernel similarity only considers the distance between two descriptors and ignores the relative position.
Specifically, $s(a_i,a_j)$ would be equal to $s(a_i,a_k)$ if $\| a_i - a_j \|_2 = \| a_i - a_k \|_2$, even though $a_j$ and $a_k$ lie on the different positions of the hyper-sphere whose center is $a_i$ and radius is $\| a_i - a_j \|_2$ or $\| a_i - a_k \|_2$.

In this paper, we propose the \textbf{linear combination weights} to measure the topological relationship between $a_i$ and its $k$NN descriptors.
Obviously our first step is to solve $k$NN descriptors for $a_i$, which are noted as $a_{ij}$ for $j=1,2,...,k$.
Then we try to linearly fit $a_i$ using $a_{ij}$ so that we minimize difference between $a_i$ and the linear combination of $a_{ij}$:
\begin{equation}
w_{ij}^a = \mathop{\arg\min}_{w_{ij}^a}  \| a_i-\sum_{j=1}^k w_{ij}^a a_{ij} \|^2
\label{eq_lcw}
\end{equation}
where linear combination weight $w_{ij}^a$ indicates topological relationship between $a_i$ and $a_{ij}$.

When $w_{ij}^a$ can be arbitrary real number, $w_{ij}^a a_{ij}$ denotes a linear space marked as ${\rm span}(a_{i1}, a_{i2},...,a_{ik})$, which is the subspace of $d$-dimensional Euclidean space, where $d$ is the length of learned descriptors and equal to $128$ in our paper.
Our purpose is to find the weights corresponding to the minimum distance between $a_i$ and ${\rm span}(a_{i1}, a_{i2},...,a_{ik})$.
However, we may have countless solutions for weights $w_{ij}^a$ when $a_i$ is in ${\rm span}(a_{i1}, a_{i2},...,a_{ik})$.
To avoid this situation, we stipulate $k$ is much smaller than $d$.

Assume ${\rm W}_i^a = [w_{i1}^a, w_{i2}^a, ... w_{ik}^a]^T \in R^{k \times 1}$ and ${\rm N}_i^a = [a_{i1}, a_{i2},...,a_{ik}] \in R^{d \times k}$. Now transform Eq.~\ref{eq_lcw} into matrix form:
\begin{equation}
{\rm W}_i^a = \mathop{\arg\min}_{{\rm W}_i^a}  \| a_i-{\rm N}_i^a {\rm W}_i^a \|^2
\end{equation}
We note that above equation is the standard \textit{Least Squares problem} so that ${\rm W}_i^a$ has the following closed-form solution:
\begin{equation}
{\rm W}_i^a = ({{\rm N}_i^a}^T {\rm N}_i^a)^{-1} {{\rm N}_i^a}^T a_i
\label{eq_Wia}
\end{equation}
When $a_{ij}$ is linearly independent with each other and $k$ is smaller than $d$, we have ${\rm rank}({{\rm N}_i^a}^T {\rm N}_i^a) = {\rm rank}({\rm N}_i^a) = k$, so that ${{\rm N}_i^a}^T {\rm N}_i^a \in R^{k \times k}$ is invertible.
There is no doubt that we can solve topology weights ${\rm W}_i^p$ for descriptor $p_i$ by the same steps.

Compared with hard weight and heat kernel similarity, our linear combination weights have the following advantages:
First, the gradient of Equation~\ref{eq_Wia} is defined everywhere, which contributes to the differential loss function for CNNs' training.
Second, no tunable parameter is required when solving our linear combination weights, which avoids the inappropriate experimental setup by human labor.
Last, every linear combination weight $w_{ij}^a$ retains the information of the whole neighborhood set, specifically, all weights would be affected and changed when a sole descriptor changes.

\subsection{Topology Measure}
\label{TM}
In previous section, we present our linear combination weights to depict the topological relationship between the center descriptor and its $k$NN descriptors.
However, the linear combination weights ${\rm W}_i^a$ and ${\rm W}_i^p$ are defined in the small neighborhood sets of $a_i$ and $p_i$, which can not be used to compare the neighborhood difference between $a_i$ and $p_i$ as a result from the misalignment of their $k$NN descriptors.

In this paper, we propose the global mapping which maps the local weights ${\rm W}_i^a$ and ${\rm W}_i^p$ to the global topology vector ${\rm T}_i^a=[t_{i1}^a, t_{i2}^a, ..., t_{in}^a]$ and ${\rm T}_i^p=[t_{i1}^p, t_{i2}^p, ..., t_{in}^p]$, where $n$ is the batch size.
The $j$-th element of ${\rm T}_i^a$ can be determined by following equation:
\begin{equation}
{t_{ij}^a}=\left\{
\begin{aligned}
&w_{i \bullet}^a, \ \ a_j \in N(a_i)\\
&0, \ \ \ \ \ otherwise\\
\end{aligned}
\right.
\label{eq_ta}
\end{equation}
Where $w_{i \bullet}^a$ is one of the elements in ${\rm W}_i^a$ and equal to linear combination weight between $a_i$ and $a_j$.
Note that  ${\rm T}_i^p$, the topology vector for descriptor $p_i$ can be established by the same way: the $j$-th element of ${\rm T}_i^p$ is equal to $w_{i \bullet}^p$ if descriptor $p_j$ is one of $k$NN descriptors of $p_i$ and equal to $0$ if not.
By above mapping, the linear combination weights with $k$ elements are transformed to the topology vectors with $n$ elements.
The topology vectors are sparse vectors because they only consist of $k$ non-zero numbers and $k$ is much smaller than $n$.

The global topology vector ${\rm T}_i^a$ or ${\rm T}_i^p$ indicates the topological relationship between center descriptor $a_i$ or $p_i$ and other descriptors within a training batch, then we define topology distance between $a_i$ and $p_i$ as the $l_1$ distance of their topology vectors:
\begin{equation}
d_T(a_i, p_i) = \frac{1}{k} \| {\rm T}_i^a - {\rm T}_i^p \|_1
\label{eq_TD}
\end{equation}
To learn the topology consistent descriptors, the topology distance between matching descriptors should be minimized during CNNs' training.

In Eq.~\ref{eq_TD}, $l1$ distance between topology vector ${\rm T}_i^a$ and ${\rm T}_i^p$ is equal to the sum of element-wise difference in ${\rm T}_i^a$ and ${\rm T}_i^p$ divided by $k$.
Firstly, when $a_j \notin N(a_i)$ and $p_j \notin N(p_i)$, $t_{ij}^a$ and $t_{ij}^p$ are both equal to $0$.
Secondly, when $a_j \notin N(a_i)$ and $p_j \in N(p_i)$, $t_{ij}^a$ is equal to $0$ and $t_{ij}^p$ not. Under this case, absolute value of $t_{ij}^p$ is encouraged to be closed to $0$ with $d_T(a_i, p_i)$ minimized, which means $p_j$ is encouraged to be far away with $p_i$ until $p_j$ is not in the neighborhood set of $p_i$.
Thirdly, when $a_j \in N(a_i)$ and $p_j \in N(p_i)$, $t_{ij}^a$ and $t_{ij}^p$ are both non-zero values and $\left| t_{ij}^a-t_{ij}^p \right|$ is minimized, which satisfies our \textbf{Assumption 2} of Section~\ref{BNC}.

Above analyses show that our topology vector depicts the neighborhood information of descriptor and we can learn the neighborhood topology consistent descriptors by minimizing the distance of topology vectors of matching descriptors.
The detailed steps to solve topology distance of the whole training batch are summarized in Algorithm~\ref{algo_topo_distance}.

\begin{algorithm}[t]
	\caption{Topology Distance}
	\label{algo_topo_distance}
	\LinesNumbered
	\KwIn{$A=\{ a_1, a_2, ..., a_n \}$ and $P=\{ p_1, p_2, ..., p_n \}$}
	\KwOut{Topology distance $d_T(a_i, p_i)$ for $i=1,2,...,n$}
	\For{$i=1:n$}
	{
	${\rm N}_i^a$ $\longleftarrow$ $[a_{i1}, a_{i2},...,a_{ik}]$ \tcp*{$k$NN descriptors of $a_i$}
	${\rm N}_i^p$ $\longleftarrow$ $[p_{i1}, p_{i2},...,p_{ik}]$ \tcp*{$k$NN descriptors of $p_i$}
	${\rm W}_i^a$ $\longleftarrow$ $({{\rm N}_i^a}^T {\rm N}_i^a)^{-1} {{\rm N}_i^a}^T a_i$; \\		
	${\rm W}_i^p$ $\longleftarrow$ $({{\rm N}_i^p}^T {\rm N}_i^p)^{-1} {{\rm N}_i^p}^T p_i$; \\
	${\rm T}_i^a$ $\longleftarrow$ ${\rm W}_i^a$ \tcp*{global mapping by Equation~\ref{eq_ta}}
	${\rm T}_i^p$ $\longleftarrow$ ${\rm W}_i^p$; \\
	$d_T(a_i, p_i)$ $\longleftarrow$ $\frac{1}{k} \| {\rm T}_i^a - {\rm T}_i^p \|_1$
	}
\end{algorithm}

\subsection{Adaptive Weighting Strategy}
\label{AWS}
Triplet loss is widely used for learning descriptors by former works~\cite{HardNet,DSM,SOSNet,ExpLoss}, which encourages the distance of negative samples is a margin larger than that of positive samples. In descriptors learning, the positive samples and negative samples indicate the matching descriptors and non-matching descriptors respectively.
The triplet loss has the following uniform format:
\begin{equation}
L_{triplet}= \frac{1}{n}\sum_{i=1}^n max(0,margin+d^+_i - d^-_i)
\label{triplet_loss}
\end{equation}
Former works define $d^+_i$ and $d^-_i$ as the Euclidean distance of matching pairs and non-matching pairs respectively, which neglect the neighborhood information of descriptors.

In this paper, we jointly minimize the topology distance and Euclidean distance of matching descriptors to learn topology consistent descriptors:
\begin{equation}
d^+_i = \lambda d_T(a_i, p_i) + (1-\lambda)d_E(a_i, p_i)
\label{eq_d+}
\end{equation}
where $d_E(a_i, p_i)$ denotes Euclidean distance between $a_i$ and $p_i$ and $\lambda$ is the hyper-parameter to balance our topology distance and Euclidean distance.

In Eq.~\ref{eq_d+}, the weight $\lambda$ is an important parameter that directly affects the performance of descriptors.
Note that $d^+_i$ with smaller $\lambda$ focuses more on the distance between individual descriptors and contributes to the more discriminative descriptors, while $d^+_i$ with a larger $\lambda$ focuses more on the neighborhood information between descriptors and contributes to the more robust descriptors.

In this paper, we employ an adaptive strategy to adjust $\lambda$ automatically during CNNs' training.
In the early training stage, the learned descriptors are not stable enough, so we should focus less on the neighborhood information.
In the later training stage, we focus more on neighborhood information to learn more robust descriptors since training CNNs using only point-to-point Euclidean distance will result in overfitting.

\textbf{Assumption 1} of Section~\ref{BNC} constrain neighborhood descriptors of $a_i$ match neighborhood descriptors of $p_i$, however, this perfect matching rarely appears as a result from the inconsistent and unstable neighborhood information of descriptors.
In this paper, we employ the number of matching pairs within two neighborhood sets $N(a_i)$ and $N(p_i)$ to measure the stability of learned descriptors: The more matching pairs denote the more stable neighborhood information of descriptors.
$N(a_i)$ and $N(p_i)$ consist of $k$NN descriptors of $a_i$ and $k$NN descriptors of $p_i$ respectively, so the maximum number of matching pairs within $N(a_i)$ and $N(p_i)$ is $k$.
We determine the value of $\lambda$ by the following formula:
\begin{equation}
\lambda = \mathop{\min} \{[ \frac{m(N(a_i),N(p_i))}{k} ]^\gamma, 0.5 \}
\label{eq_gamma}
\end{equation}
where $m(N(a_i),N(p_i))$ denotes the number of matching pairs and $\gamma$ is the tunable parameter.
By this method can we enable the adaptive and automatic adjustment of $\lambda$ in Eq.~\ref{eq_d+} during CNNs' training.

\section{Experiments}
\label{experiments}
The main contribution of our work is to propose the topology measure besides Euclidean distance to encourage the similar topology of matching descriptors. 
In the ablation studies of Section\ref{Ablation_Studies}, we choose DSM~\cite{DSM} as our baseline, which is the state-of-the-art method of learning descriptors by triplet loss.
And in the extensive experiments of ~\ref{Extensive_Experiments}, we test our method on the basis of both HardNet~\cite{HardNet} and DSM~\cite{DSM} to verify the generalization of our method, where HardNet first introduces triplet loss into learning descriptors.

To validate the performance of our topology consistence descriptors \textbf{TCDesc}, we conduct our experiments in four benchmarks: UBC PhotoTourism~\cite{UBC}, HPatches~\cite{HPatches}, W1BS dataset~\cite{WxBS} and Oxford dataset~\cite{Oxford}.
UBC PhotoTourism~\cite{UBC} is currently the largest and the most widely used local image patches matching dataset, which consists of three subsets(\textit{Liberty}, \textit{Notredame} and \textit{Yosemite}) with more than 400k image patches.
HPatches~\cite{HPatches} presents the more complicated and more comprehensive three tasks to evaluate descriptors: \textit{Patch Verification}, \textit{Image Matching}, and \textit{Patch Retrieval}.
W1BS dataset~\cite{WxBS} consists of 40 image pairs and provides more challenging tasks with several nuisance factors to explore the performance of descriptors in extreme conditions.
Oxford dataset~\cite{Oxford} presents the real image matching scenarios, in which matching score is taken to evaluate the performance of learned descriptors.

\subsection{Implementations}
We use the same configuration as former works to guarantee the improvement of experimental results attributes to our novel topology measure.
We use the CNN architecture proposed in L2-Net~\cite{L2Net} with seven convolutional layers and a Local Response Normalization layer.
We only train our network on benchmark UBC PhotoTourism and then test other three benchmarks using the trained model.
The size of image patches in UBC PhotoTourism is $64 \times 64$, then we downsample each patch to size of $32 \times 32$, which is required by of L2-Net.
We conduct data augmentation as DSM~\cite{DSM} to flip or rotate image patches randomly.
To accord with HardNet~\cite{HardNet} and DSM~\cite{DSM}, we set the training batch size $n$ to be 1024.
We train our network for $150k$ iterations using Stochastic Gradient Descent(SGD) with momentum $0.9$ and weight decay $10^{-4}$, and the learning rate is decayed linearly from $0.1$ to $0$.

\subsection{Ablation Studies}
\label{Ablation_Studies}
In this section, three ablation experiments are conducted to verify the effectiveness of our method: We first explore the performances of learned descriptors under different $k$ and $\gamma$ in Section~\ref{impact_k}, then compare three categories of topology weights (hard weights, heat kernel similarity and our linear combination weights) in Section~\ref{Comparison}, last verify the validity of adaptive weighting strategy in Section~\ref{validity}.

\begin{figure*}[ht]
	\centering
	\subfigure[Patch Verification]{\includegraphics[width=0.32\textwidth]{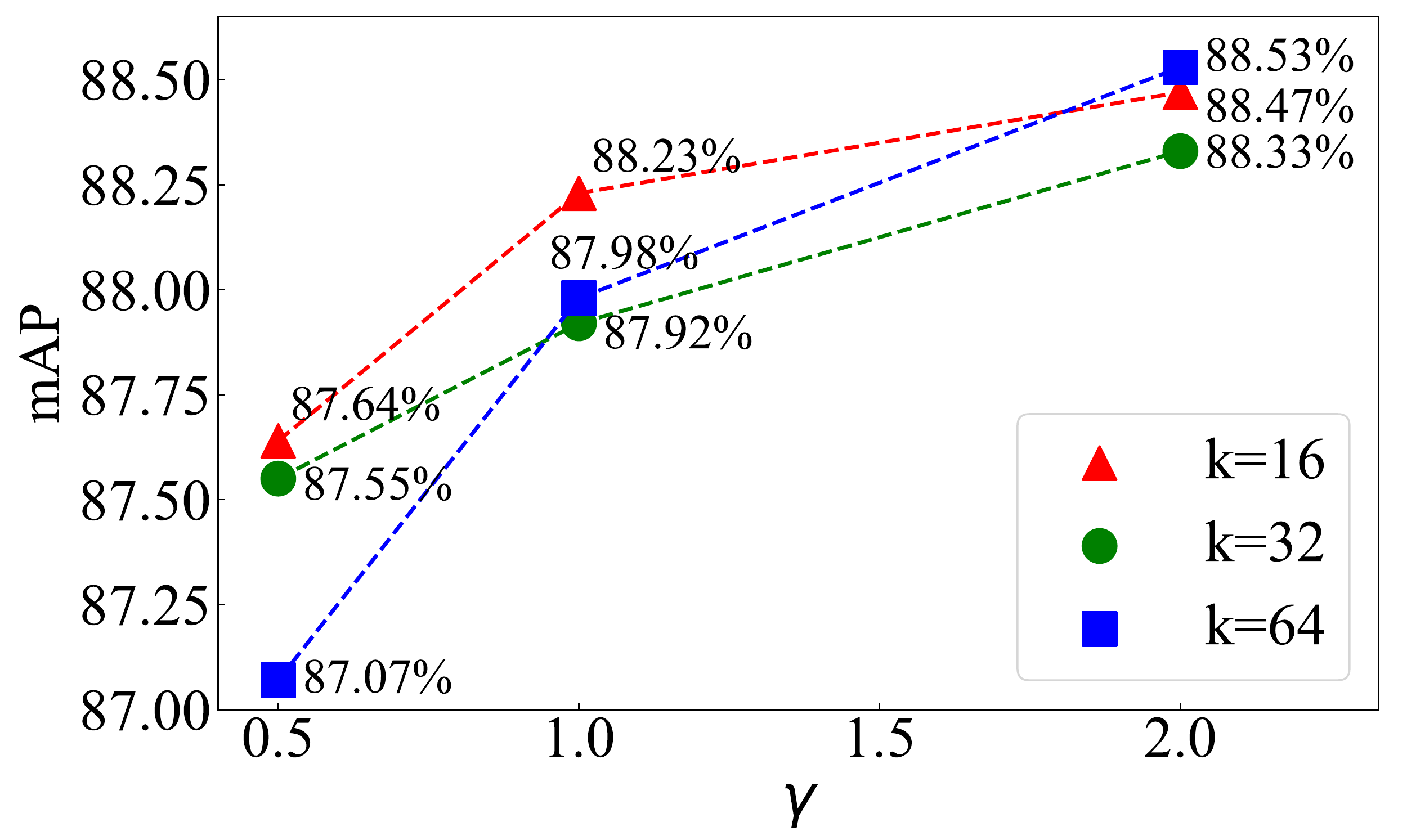}}
	\subfigure[Image Matching]{\includegraphics[width=0.32\textwidth]{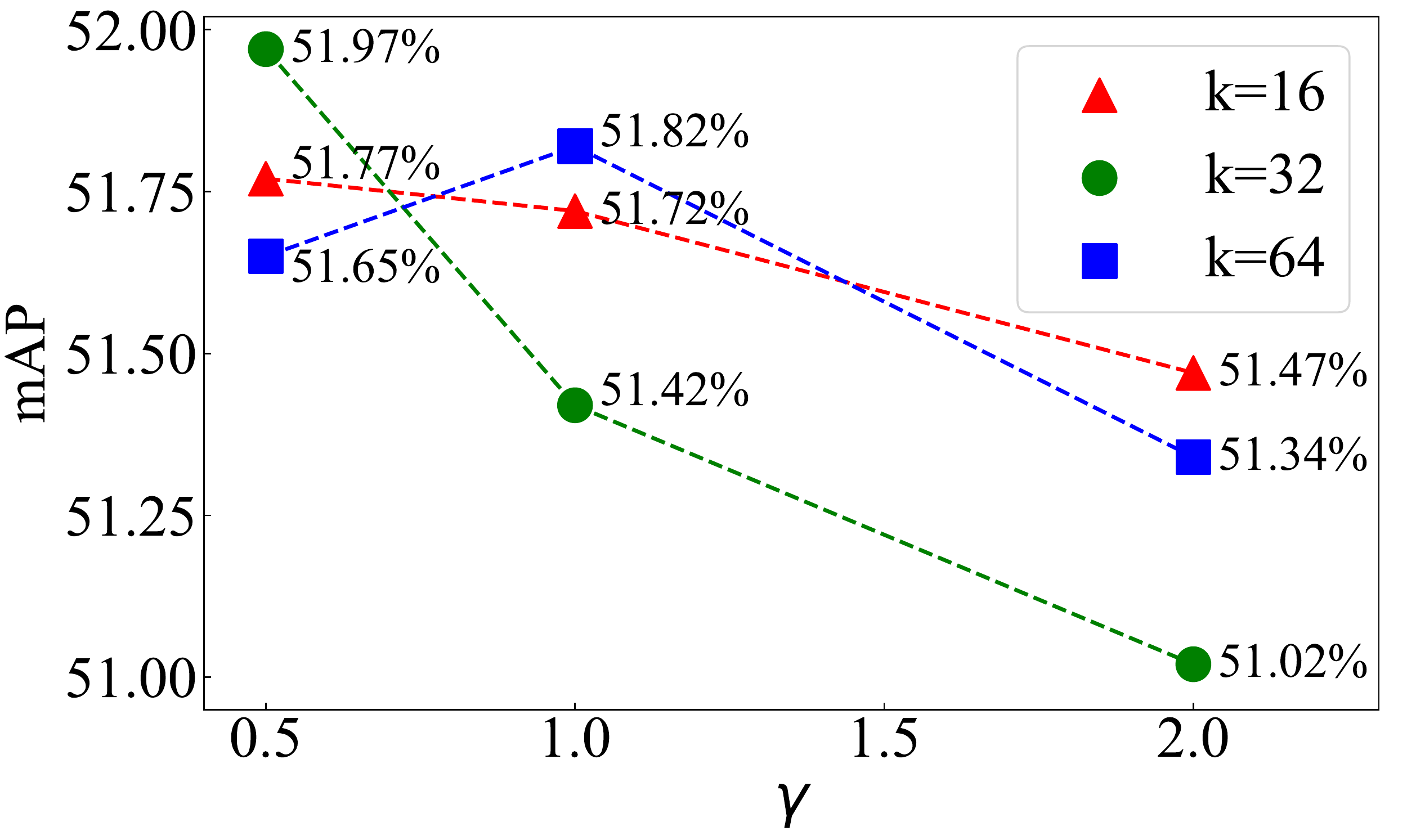}}
	\subfigure[Patch Retrieval]{\includegraphics[width=0.32\textwidth]{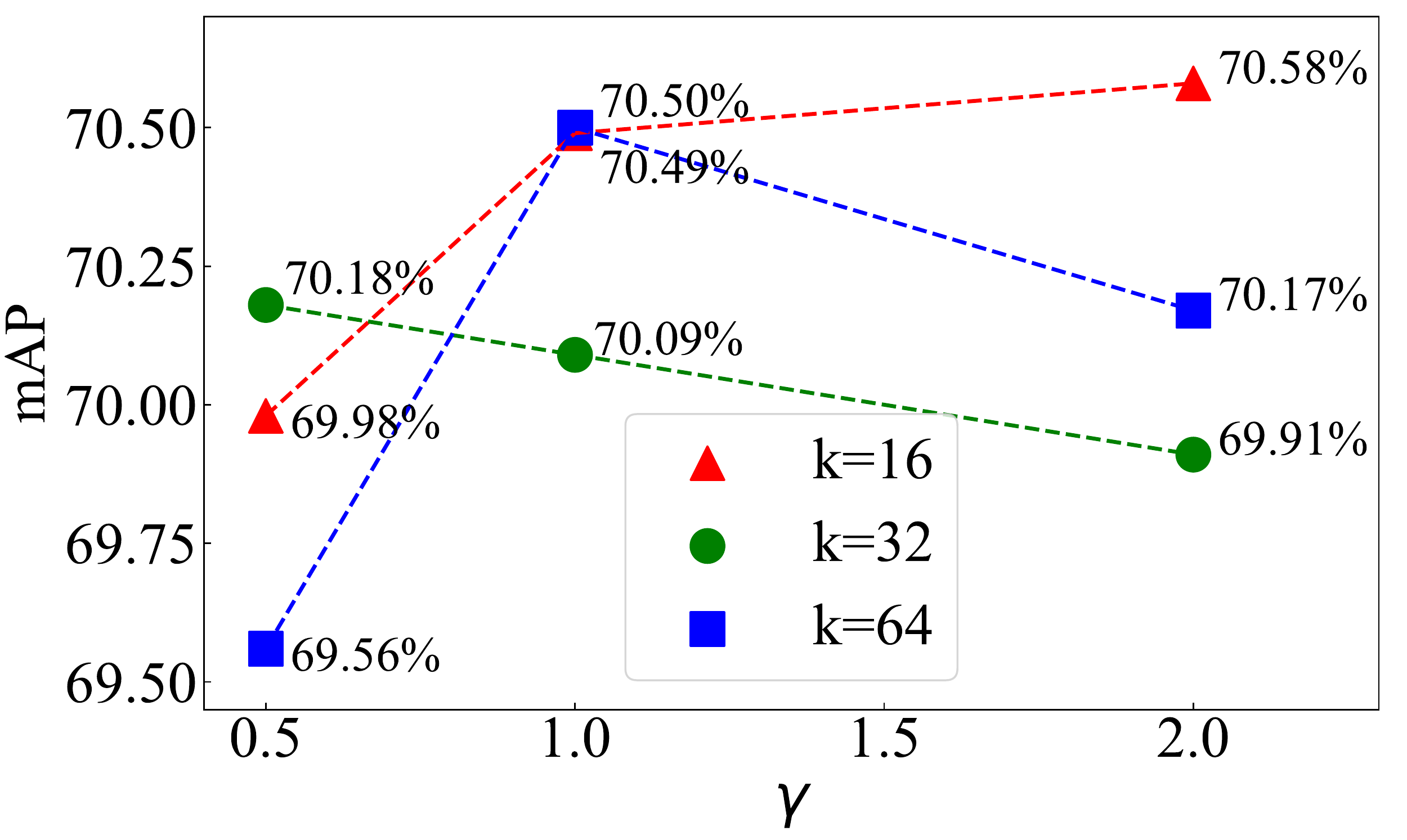}}
	\caption{Performance of descriptors under different $\gamma$ and $k$ on HPatches benchmark. In three tasks, the lager mAP denotes the better performance. We found that the lager $gamma$ contributes better performance of descriptors on task \textit{Patch Verification} and worse performance of descriptors on task \textit{Image Matching}.}
	\label{fig_gamma_k}
\end{figure*}

\subsubsection{Impact of $k$ and $\gamma$}
\label{impact_k}
The hyper-parameter $k$ denotes the number of descriptors in neighborhood sets of center descriptor $a_i$ or $p_i$, while the lager $k$ means we take more neighborhood information for CNNs' training, however, the lager $k$ also results in lager computation and countless solutions for our linear combination weights. In this paper, we explore three values of $k$: $16$, $32$ and $64$.
In Eq.~\ref{eq_d+}, parameter $\lambda$ denotes the weighting coefficient to fuse topology distance and Euclidean distance, which is decided by $\gamma$ of Eq.~\ref{eq_gamma}.
In this paper, we explore three values of $\gamma$ as well: $0.5$, $1$ and $2$.

We first conduct our experiment on UBC PhotoTourism benchmark~\cite{UBC}.
UBC PhotoTourism~\cite{UBC} is the first large benchmark of learning descriptors from image patches which consists of more than 400k image patches extracted from large 3D reconstruction scenes.
UBC PhotoTourism consists of three subsets: \textit{Liberty}, \textit{Notredame} and \textit{Yosemite}.
Usually we train on one sbuset and test on other two subsets.
The false positive rate at 95\% recall (FPR95) is employed to evaluate the performance of learned descriptors, where the lower FPR95 indicates the better performance.

We choose DSM~\cite{DSM} as our baseline, which is the current state-of-the-art model of learning descriptors by triplet loss.
In this experiment, we choose to train our model using \textit{Liberty} subset and test on another two subsets, and the experimental results are presented in Table.~\ref{table_UBC_kgamma}.
As can be seen, descriptors learned by our method under different $\gamma$ and $k$ all outperform or perform equally than DSM~\cite{DSM}, and we get our lowest FPR95 when $\gamma$ is equal to $1.0$ and k is equal to $32$.

In Table.~\ref{table_UBC_kgamma}, we found that descriptors learned under different $k$ or $\gamma$ may result the similar or even the same performance, such as case '$k=16, \ \gamma=1.0$' and case '$k=64, \ \gamma=1.0$'.
We then conduct our experiment on HPatches benchmark~\cite{HPatches}. 
HPatches benchmark~\cite{HPatches} consists of 116 sequences where the main nuisance factor of 57 sequences is illumination and that of 59 sequences is viewpoint.
Compared with UBC PhotoTourism benchmark, HPatches benchmark~\cite{HPatches} provides more diverse data samples and more sophisticated tasks.
HPatches~\cite{HPatches} defines three tasks to evaluate descriptors: \textit{Patch Verification}, \textit{Image Matching}, and \textit{Patch Retrieval}, and  each task is categorized as "Easy", "Hard" or "Tough" according to the amount of geometric noise or changes in viewpoint and light illumination.
The mean average precision(mAP) is adopted to evaluate descriptors and the higher mAP indicates the better performance.

We use model trained on subsets \textit{Liberty} of UBC PhotoTourism benchmark to generate descriptors from image patches of HPatches.
The experimental results are presented in Fig.~\ref{fig_gamma_k}.
As can be seen, the lager $\lambda$ contributes better performance of descriptors on task \textit{Patch Verification}, while the smaller $\lambda$ contributes better performance of descriptors on task \textit{Image Matching}.
Otherwise, performance of descriptors is not relevant to the value of $k$ in three tasks of HPatches benchmark.

With the comprehensive consideration on both UBC PhotoTourism benchmark and HPatches benchmark, we determine the optimal value of $k$ and $\gamma$: $k=16$ and $\gamma=1.0$.
Under this setup, the mean FPR95 on UBC PhotoTourism benchmark is $0.85\%$ and mAPs on three tasks of HPatches benchmark are $88.23\%$, $51.72\%$ and $70.49\%$ respectively.

\begin{table}[ht]
	\centering
	\caption{Performance of descriptors under different $\gamma$ and $k$ on UBC PhotoTourism benchmark~\cite{UBC}. Numbers shown are FPR95(\%), while the lower FPR95 indicates the better performance. We choose DSM~\cite{DSM} as our baseline, which is the state-of-the-art model of learning descriptors by triplet loss.}
	\begin{tabular}{cccccc}
		\hline
		\multirow{2}{*}{$\gamma$} & \multirow{2}{*}{$k$} & train & \multicolumn{2}{c}{Liberty} & \multirow{2}{*}{mean} \\ \cline{4-5}
		&                    & test  & Notredime     & Yosemite    &                       \\ \hline
		\multicolumn{2}{c}{DSM~\cite{DSM}}                   &       & 0.39          & 1.51        & 0.95                  \\ \hline
		\multirow{3}{*}{0.5}   & 16                 &       & 0.36          & 1.28        & 0.82                  \\
		& 32                 &       & 0.40          & 1.37        & 0.89                  \\
		& 64                 &       & 0.39          & 1.47        & 0.93                  \\ \hline
		\multirow{3}{*}{1.0}   & 16                 &       & 0.37          & 1.33        & 0.85                  \\
		& 32                 &       & 0.36          & \textbf{1.26}        & \textbf{0.81}                  \\
		& 64                 &       & \textbf{0.32}          & 1.37        & 0.85                  \\ \hline
		\multirow{3}{*}{2.0}   & 16                 &       & 0.37          & 1.43        & 0.90                  \\
		& 32                 &       & 0.36          & 1.54        & 0.95                  \\
		& 64                 &       & 0.35          & 1.37        & 0.86                  \\ \hline
	\end{tabular}
	\label{table_UBC_kgamma}
\end{table}

\begin{table*}[ht]
	\centering
	\caption{Patch verification performance on the UBC PhotoTourism benchmark. Numbers shown are FPR95(\%), while the lower FPR95 indicates the better performance of learned descriptors. Plus "+" denotes training with data augmentation. We test our method on the basis of HardNet~\cite{HardNet} and DSM~\cite{DSM}, which is noted as TCDesc-HN and TCDesc-DSM respectively.}
	\begin{tabular}{cccccccccc}
		\hline
		\multirow{2}{*}{Descriptors} & \multirow{2}{*}{Length} & Train & Notredame     & Yosemite    & Liberty       & Yosemite      & Liberty      & Notredame     & \multirow{2}{*}{Mean} \\ \cline{4-9}
		&                         & Test  & \multicolumn{2}{c}{Liberty} & \multicolumn{2}{c}{Notredame} & \multicolumn{2}{c}{Yosemite} &                       \\ \hline
		SIFT~\cite{SIFT}                         & 128                     &       & \multicolumn{2}{c}{29.84}   & \multicolumn{2}{c}{22.53}     & \multicolumn{2}{c}{27.29}    & 26.55                 \\
		DeepDesc~\cite{DeepDesc}                     & 128                     &       & \multicolumn{2}{c}{10.9}    & \multicolumn{2}{c}{4.40}      & \multicolumn{2}{c}{5.69}     & 7.0                   \\
		L2-Net+~\cite{L2Net}                      & 128                     &       & 2.36          & 4.70        & 0.72          & 1.29          & 2.51         & 1.71          & 2.23                  \\
		CS L2-Net+~\cite{L2Net}                   & 256                     &       & 2.55          & 4.24        & 0.87          & 1.39          & 3.81         & 2.84          & 2.61                  \\
		HardNet~\cite{HardNet}                     & 128                     &       & 1.47          & 2.67        & 0.62          & 0.88          & 2.14         & 1.65          & 1.57                  \\
		HardNet+~\cite{HardNet}                     & 128                     &       & 1.49          & 2.51        & 0.53          & 0.78          & 1.96         & 1.84          & 1.51                  \\
		DOAP+~\cite{DOAP}                        & 128                     &       & 1.54          & 2.62        & 0.43          & 0.87          & 2.00         & 1.21          & 1.45                  \\
		DOAP-ST+~\cite{DOAP,STN}                     & 128                     &       & 1.47          & 2.29        & 0.39          & 0.78          & 1.98         & 1.35          & 1.38                  \\
		ESE~\cite{ESE}                          & 128                     &       & 1.14          & 2.16        & 0.42          & 0.73          & 2.18         & 1.51          & 1.36                  \\
		SOSNet~\cite{SOSNet}                       & 128                     &       & 1.25          & 2.84        & 0.58          & 0.87          & 1.95         & 1.25          & 1.46                  \\
		Exp-TLoss~\cite{ExpLoss}                    & 128                     &       & 1.16          & 2.01        & 0.47          & 0.67          & \textbf{1.32}         & \textbf{1.10}          & 1.12                  \\
		DSM+~\cite{DSM}                          & 128                     &       & 1.21          & 2.01        & 0.39          & 0.68          & 1.51         & 1.29          & 1.18                  \\ 
		TCDesc-HN+                          & 128                     &       & 1.29          & 2.20        & 0.41          & 0.69          & 1.44         & 1.29          & 1.22                  \\
		TCDesc-DSM+                          & 128                     &       & \textbf{1.12}           & \textbf{1.90}        & \textbf{0.37}          & \textbf{0.63}          & 1.33         & 1.14          & \textbf{1.08}                  \\ \hline
	\end{tabular}  \label{UBCbenchmark}
\end{table*}

\begin{table}[ht]
	\centering
	\caption{Performance of descriptors learned by different topology weights on UBC PhotoTourism benchmark~\cite{UBC}. Numbers shown are FPR95(\%), while the lower FPR95 indicates the better performance. Our linear combination weights works well on hard task (testing Yosemite subset).}
	\begin{tabular}{cccccc}
		\hline
		\multicolumn{2}{c}{\multirow{2}{*}{weight}}                                                & train & \multicolumn{2}{c}{Liberty} & \multirow{2}{*}{mean} \\ \cline{4-5}
		\multicolumn{2}{c}{}                                                                       & test  & Notredime     & Yosemite    &                       \\ \hline
		\multicolumn{2}{c}{ours}                                                                   &       & 0.37          & \textbf{1.33}        & \textbf{0.85}                  \\ \hline
		\multicolumn{2}{c}{hard weight}                                                            &       & \textbf{0.36}              & 1.42            & 0.89                      \\ \hline
		\multirow{5}{*}{\begin{tabular}[c]{@{}c@{}}heat kernel\\ similarity\end{tabular}} & t=0.1  &       & \textbf{0.36}          & 1.37        & 0.87                  \\
		& t=0.5  &       & 0.37          & 1.36        & 0.87                  \\
		& t=1.0  &       & 0.37          & 1.39        & 0.88                  \\
		& t=5.0  &       & 0.37          & 1.44        & 0.91                  \\
		& t=10.0 &       & 0.40          & 1.45        & 0.93                      \\ \hline
	\end{tabular}
	\label{table_weights}
\end{table}

\begin{table}[ht]
	\centering
	\caption{Performance of descriptors learned by different topology weights on HPatches benchmark~\cite{HPatches}. Numbers shown are mAP(\%), while the lager mAP indicates the better performance.}
	\begin{tabular}{cccccc}
		\hline
		\multicolumn{2}{c}{\multirow{2}{*}{weight}}                                                & \multicolumn{3}{c}{task}           & \multirow{2}{*}{mean} \\ \cline{3-5}
		\multicolumn{2}{c}{}                                                                       & Verfication & Matching & Retrieval &                       \\ \hline
		\multicolumn{2}{c}{ours}                                                                   & 88.23     & 51.72  & \textbf{70.49}   & \textbf{70.15}                  \\ \hline
		\multicolumn{2}{c}{hard weight}                                                            & 88.01        & 51.45     & 69.89      & 69.78                  \\ \hline
		\multirow{5}{*}{\begin{tabular}[c]{@{}c@{}}heat kernel\\ similarity\end{tabular}} & t=0.1  & 88.26        & 51.52     & 70.23      & 70.00                  \\
		& t=0.5  & 88.17        & \textbf{51.73}     & 70.26      & 70.05                  \\
		& t=1.0  & 88.13        & 51.65     & 70.43      & 70.07                  \\
		& t=5.0  & \textbf{88.24}        & 51.62     & 70.31      & 70.06                  \\
		& t=10.0 & 88.05        & 51.61     & 70.16      & 69.94                  \\ \hline
	\end{tabular}
	\label{table_HPatches_weights}
\end{table}

\subsubsection{Comparison of three categories of topology weights}
\label{Comparison}
In Section~\ref{TW} we mentioned that the hard weights and heat kernel similarity are widely adopted to measure the topological relationship between two samples and then we figure out their drawbacks.
In this section we compare our linear combination weights and the former two topology weights on both UBC PhotoTourism benchmark and HPatches benchmark.
For the fair comparison, we set $k$ to be $16$ and set $\gamma$ to be $1.0$ in all experiments of this section.

The hard weights of Equation~\ref{eq_binary} is not differential, which brings the obstacle to optimize the loss function.
So we choose a differential proxy to approximate the hard weights:
\begin{equation}
h(a_i,a_j)=\left\{
\begin{aligned}
&exp\{- \frac{\| a_i - a_j \|_2}{10^3} \}, \ \ a_j \in N(a_i)\\
&0, \ \ \ \ \ \ \ \ \ \ \ \ \ \ \ \ \ \ \ \ \ \ \ \ otherwise\\
\end{aligned}
\right.
\end{equation}
In above equation, $a_i$ and $a_j$ are both unit-length vector so that $\frac{\| a_i - a_j \|_2}{10^3}$ is a very small number and $exp\{- \frac{\| a_i - a_j \|_2}{10^3} \}$ is approximately equal to $1$.

Otherwise, heat kernel similarity of Eq.~\ref{eq_hks} consists of a hyper-parameter $t$, where the lager $t$ results the lager similarity score.
In this section, we test five values for $t$: $0.1$, $0.5$, $1.0$, $5.0$ and $10.0$.
We present the experimental results in Table~\ref{table_weights}.
We conclude that our linear combination weights work well on hard task (testing \textit{Yosemite} subset): Descriptors learned by our linear combination weights achieve the lowest FPR95 $1.33\%$.

To further verify the effectiveness of our linear combination weights, we compare the performance of descriptors learned by different weights on HPatches benchmark~\cite{HPatches}.
As shown in Table~\ref{table_HPatches_weights}, though performances of descriptors learned by different weights are close to each other, our method still outperforms other two weights.

\begin{table}[t]
	\centering
	\caption{Comparison of our adaptive weighting strategy and fixed $\lambda$. Numbers shown are FPR95(\%), while the lower FPR95 indicates the better performance. Descriptors learned by our adaptive weighting strategy outperforms than that learned by any fixed $\lambda$.}
	\begin{tabular}{cccccc}
		\hline
		\multicolumn{2}{c}{\multirow{2}{*}{strategy}}                                             & train & \multicolumn{2}{c}{Liberty} & \multirow{2}{*}{mean} \\ \cline{4-5}
		\multicolumn{2}{c}{}                                                                      & test  & Notredime     & Yosemite    &                       \\ \hline
		\multicolumn{2}{c}{ours}                                                                  &       & 0.37          & \textbf{1.33}        &  \textbf{0.85}                  \\ \hline
		\multirow{6}{*}{\begin{tabular}[c]{@{}c@{}}no adaptive\\ weighting\end{tabular}} & $\lambda$=0.05 &   & 0.39          & 1.42        & 0.91                  \\
		& $\lambda$=0.1  &   & 0.37          & 1.42        & 0.90                  \\
		& $\lambda$=0.2  &   &  \textbf{0.35}          & 1.50        & 0.93                  \\
		& $\lambda$=0.3  &   & 0.38          & 1.35        & 0.92                  \\
		& $\lambda$=0.4  &   & 0.39          & 1.37        & 0.88                  \\
		& $\lambda$=0.5  &   & 0.46          & 1.34        & 0.90                  \\ \hline
	\end{tabular}
	\label{table_diff_lambda}
\end{table}

\subsubsection{Validity of adaptive weighting strategy}
\label{validity}
In section ~\ref{AWS} we propose the adaptive weighting strategy to adjust the weight $\lambda$ of Eq~\ref{eq_d+} automatically.
In this section, we explore and compare the performance of descriptors learned by our adaptive weighting strategy and fixed $\lambda$ on UBC PhotoTourism benchmark~\cite{UBC}.
We test six values for $\lambda$: $0.05$,  $0.1$, $0.2$, $0.3$, $0.4$ and $0.5$.

For the fair comparison, we conduct our experiments using the same settings as Section~\ref{impact_k} except the value of $\lambda$.
As shown in Table~\ref{table_diff_lambda}, descriptors learned by our adaptive weighting strategy outperforms than that learned by any fixed $\lambda$: we achieve the lowest mean FPR95 $0.85\%$.

\begin{figure*}[ht]
	\centering
	\subfigure{\includegraphics[width=0.32\textwidth]{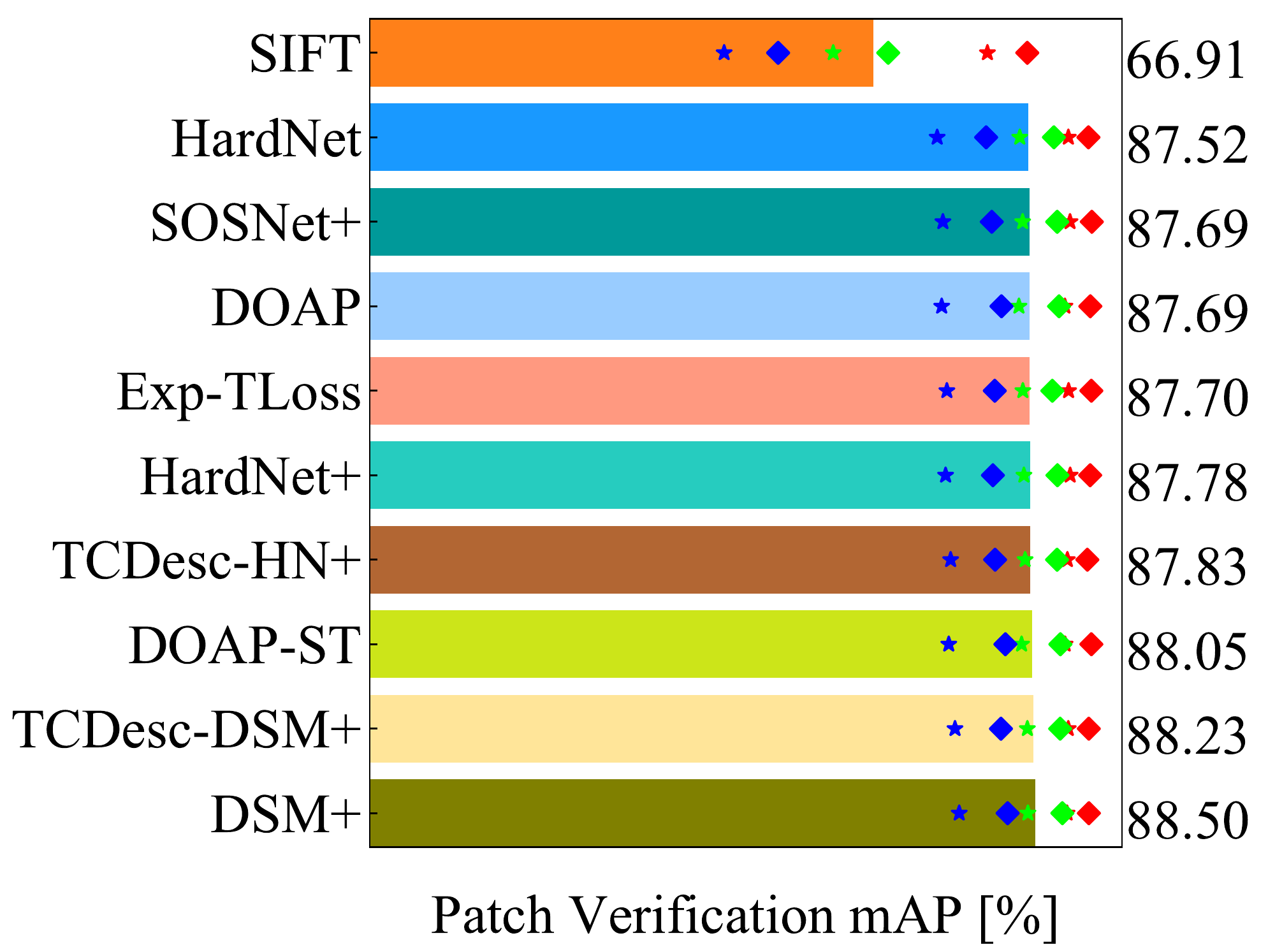}}
	\subfigure{\includegraphics[width=0.32\textwidth]{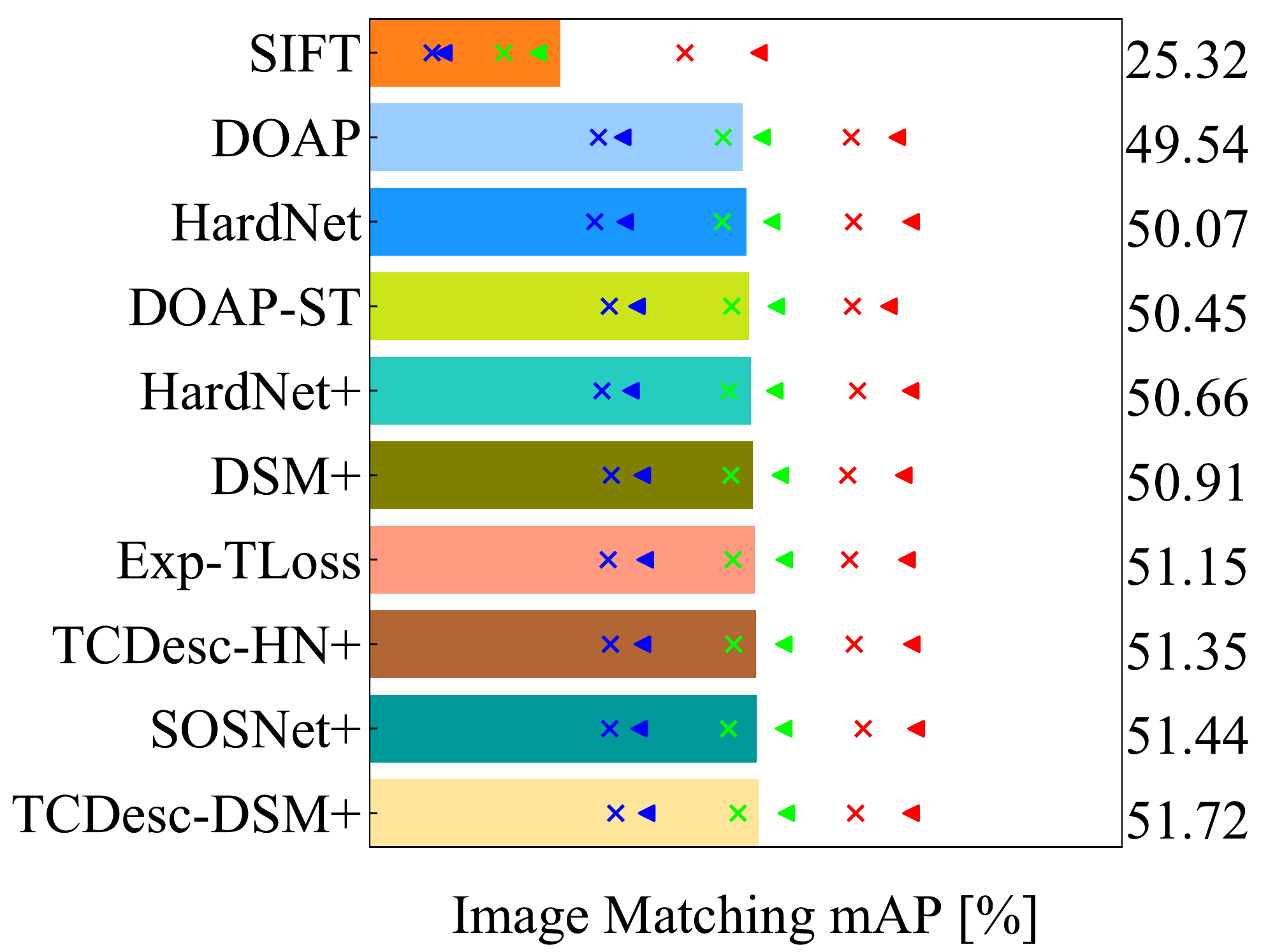}}
	\subfigure{\includegraphics[width=0.32\textwidth]{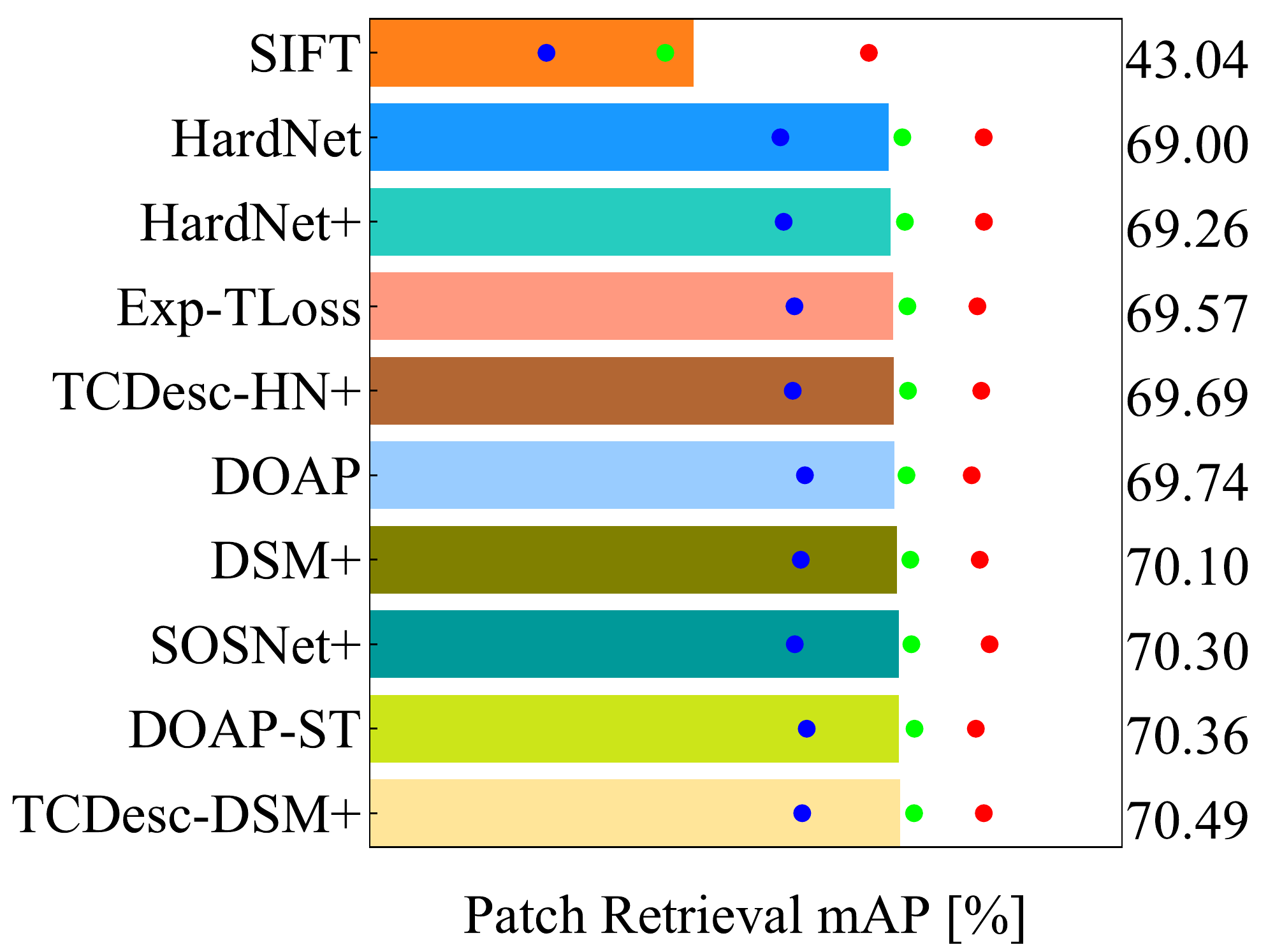}}
	\caption{Performance of descriptors on HPatches benchmark~\cite{HPatches}. In these three figures, colors of markers indicate the difficulty level of tasks: easy (red), hard (green), tough (blue). In the left figure, DIFFSEQ($\diamond$) and SAMESEQ($\star$) represent the source of negative examples in verification task. In the middle figure,  ILLUM ($\times$) and VIEWPT ($\triangleleft$) indicate the influence of illumination and viewpoint changes in matching task. All the descriptors are generated by the model trained on subsets \textit{Liberty} of UBC PhotoTourism benchmark.}
	\label{HPatches}
\end{figure*}

\begin{figure*}[ht]
	\centering
	\subfigure[Appearance]{\includegraphics[width=0.30\textwidth]{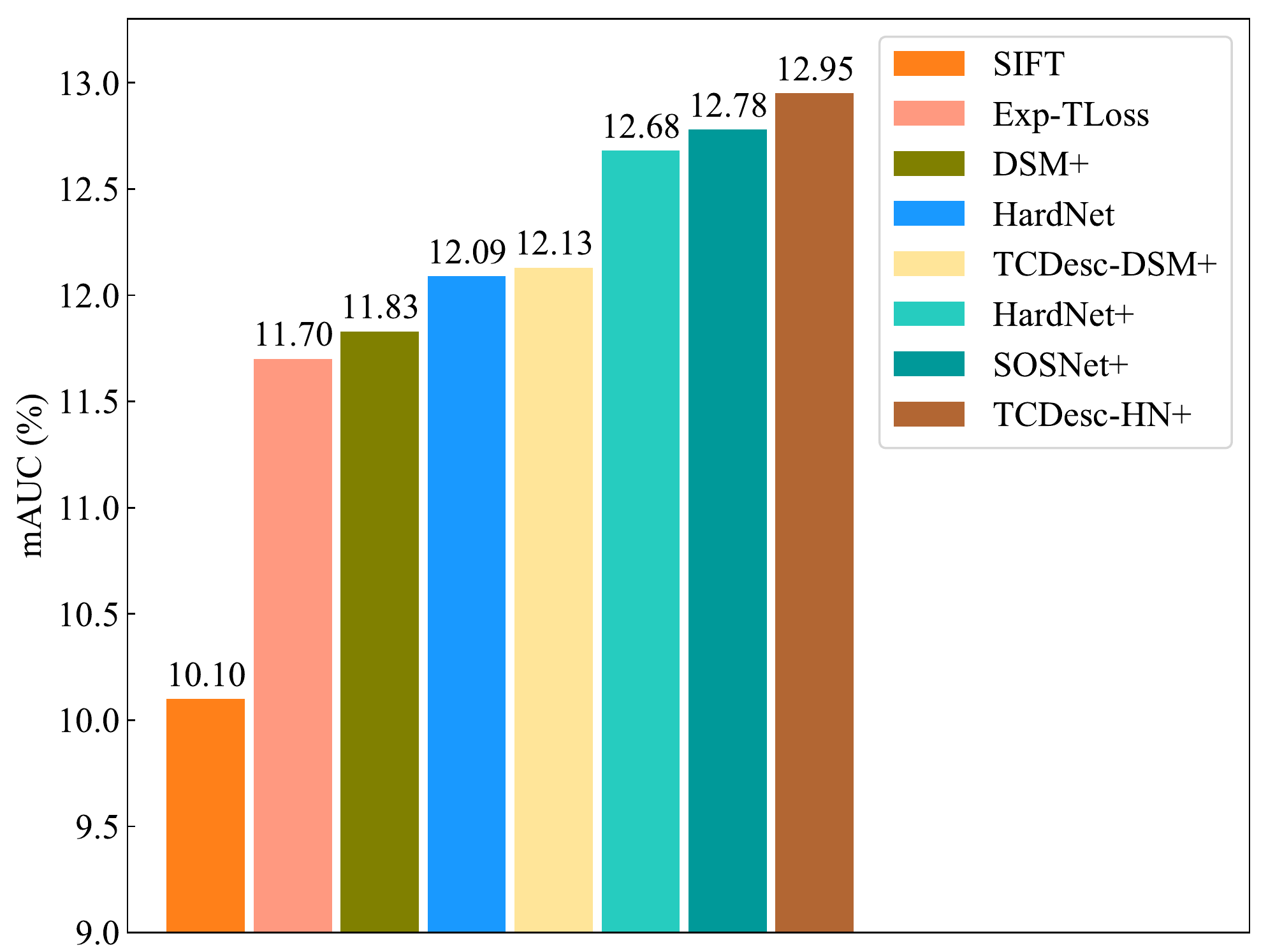}}
	\subfigure[Geometry]{\includegraphics[width=0.30\textwidth]{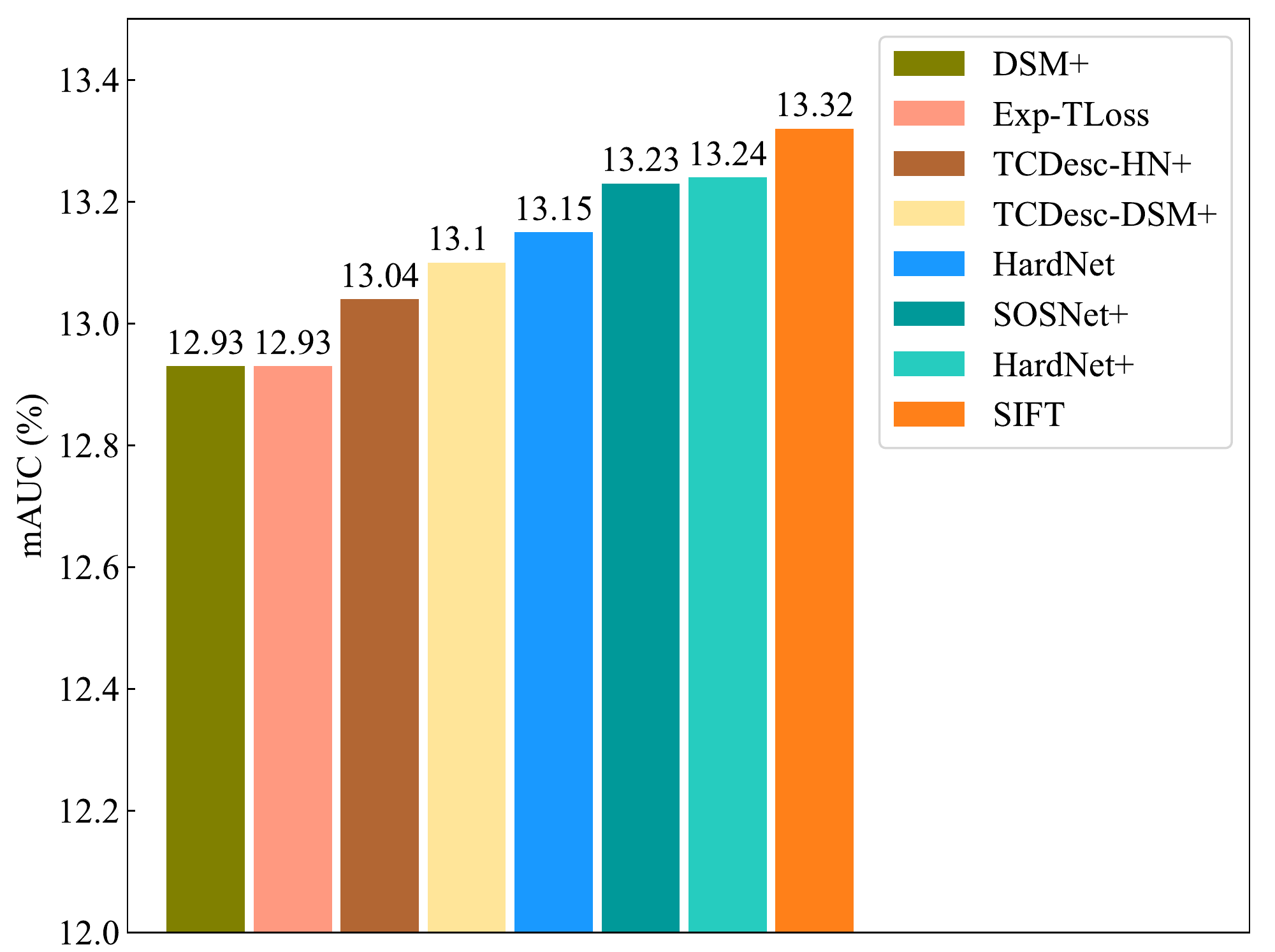}}
	\subfigure[Illumination]{\includegraphics[width=0.30\textwidth]{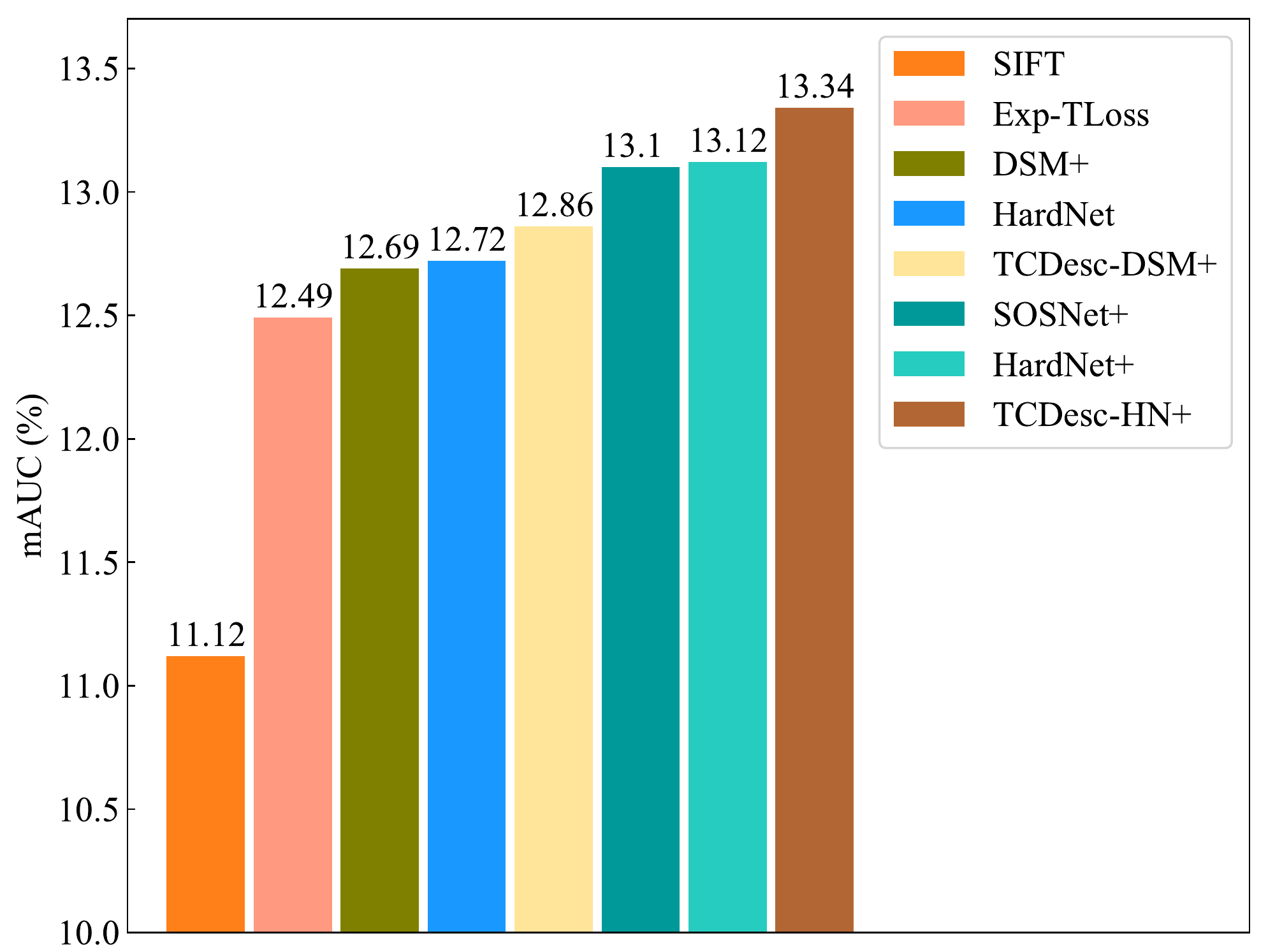}}
	\subfigure[Sensor]{\includegraphics[width=0.30\textwidth]{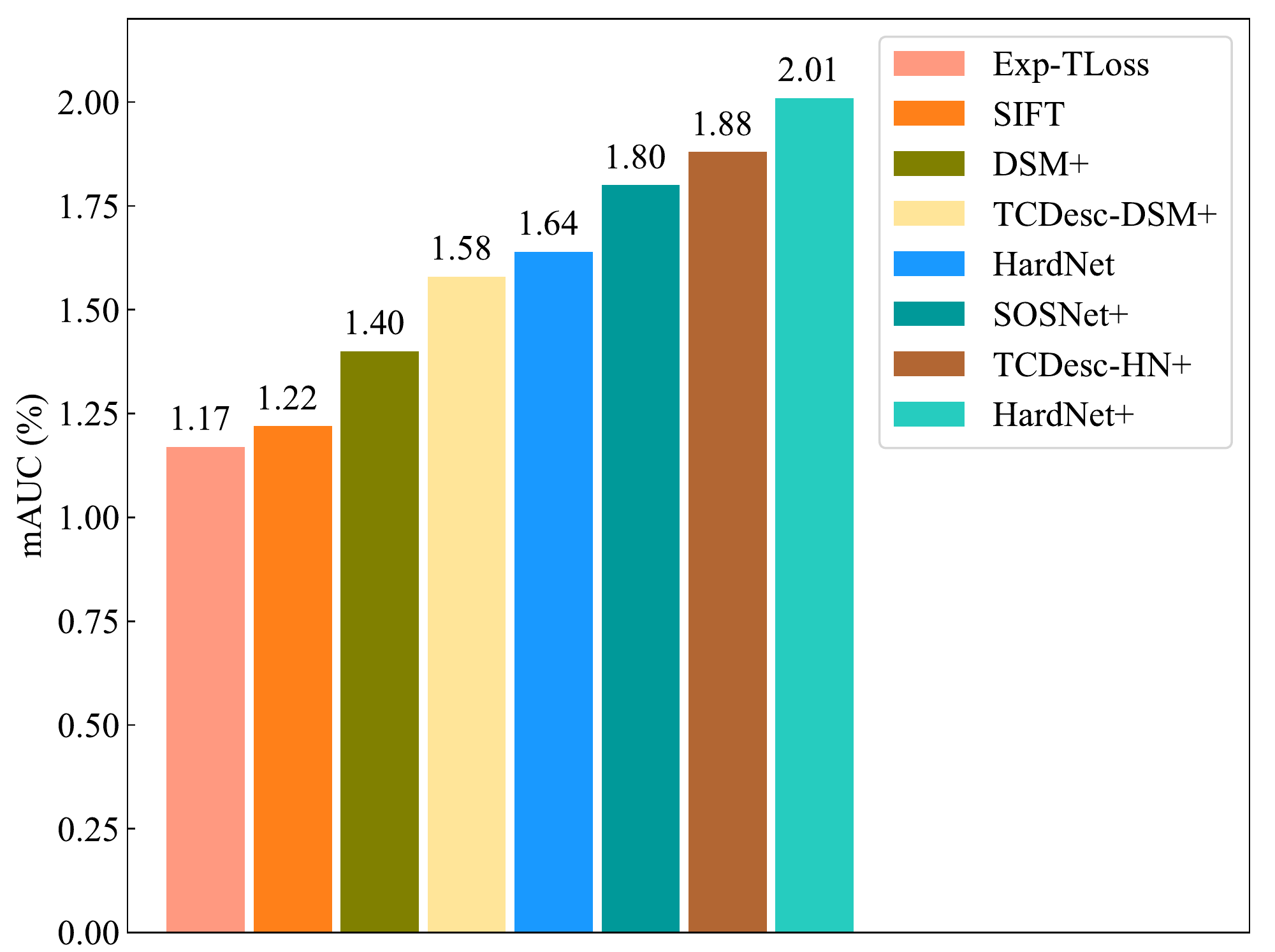}}
	\subfigure[Map2Photo]{\includegraphics[width=0.30\textwidth]{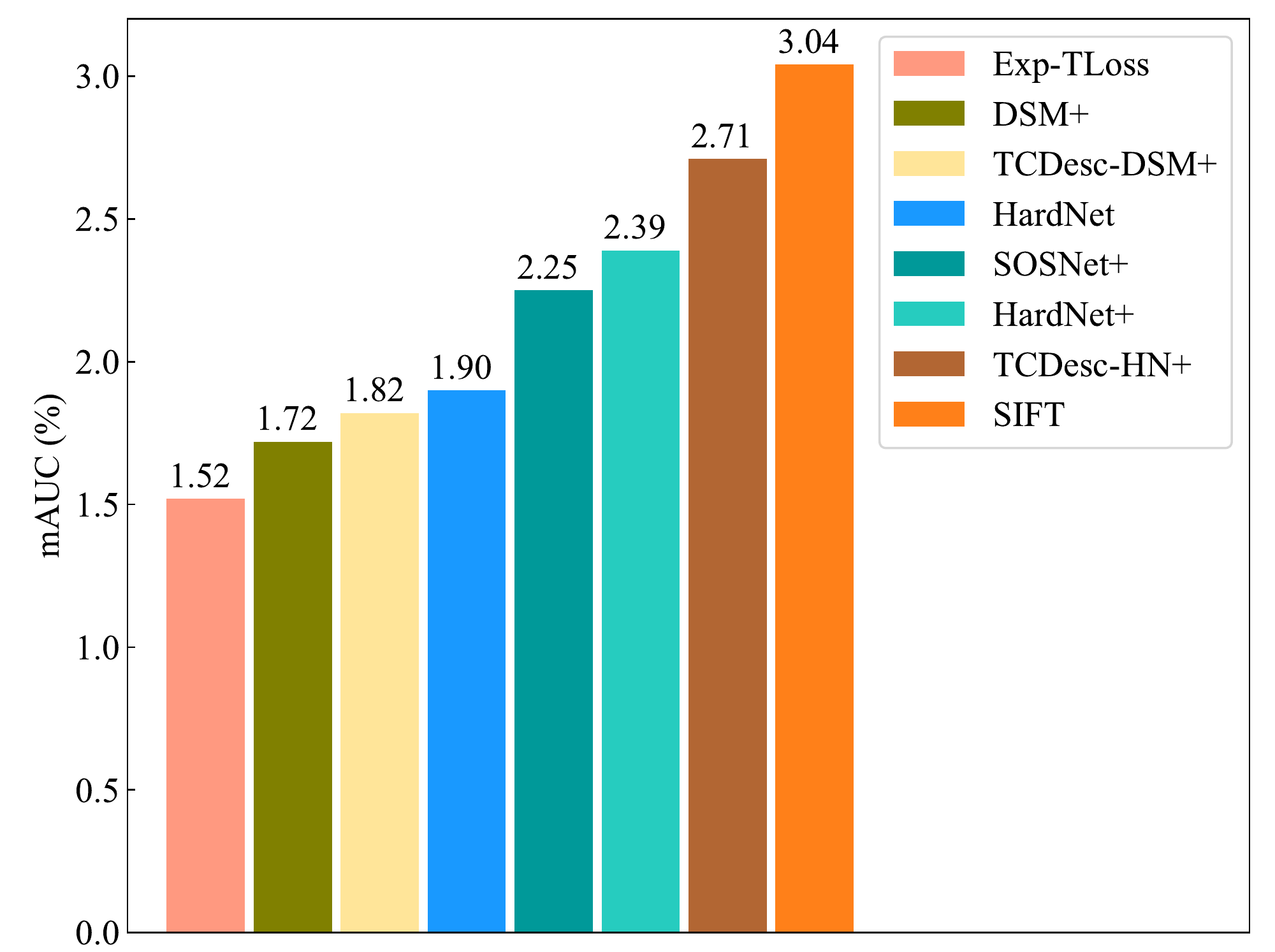}}
	\subfigure[mean]{\includegraphics[width=0.30\textwidth]{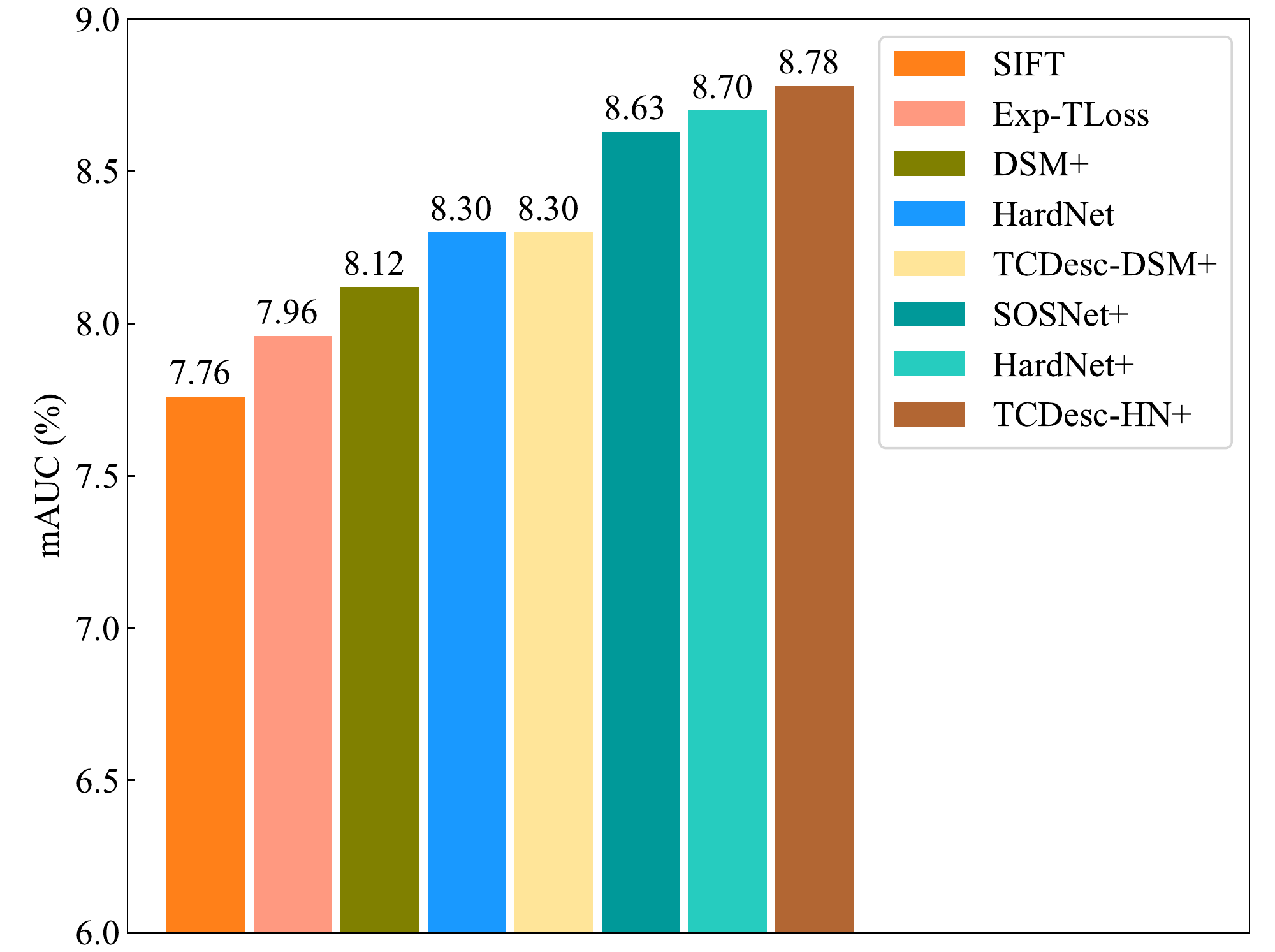}}
	\caption{Performance of descriptors on W1BS benchmark~\cite{WxBS}. W1BS benchmark consists of five subtasks divided by the main nuisance factor: Appearance, Geometry, Illumination, Sensor, and Map2Photo. Numbers shown in figures are mAUC and the lager mAUC denotes the better performance.}
	\label{wxbs}
\end{figure*}

\begin{table*}[ht]
	\centering
	\caption{Performance of descriptors learned by different algorithms on Oxford dataset~\cite{Oxford}. Numbers shown in the table denote the matching scores (\%) and lager matching score means better performance of the learned descriptors.}
	\begin{tabular}{cccccccccc}
		\hline
		\multirow{2}{*}{methods} & \multicolumn{8}{c}{subset}                                & \multirow{2}{*}{mean} \\ \cline{2-9}
		& leuven & graf & boat & bark & wall & ubc  & trees & bikes &                        \\ \hline
		HardNet~\cite{HardNet}                  & 52.04   & 30.20 & 31.88 & 8.68 & 34.80 & 65.24 & 20.44  & 39.36  & 35.33                   \\
		HardNet+~\cite{HardNet}                 & 52.56   & 30.84 & 33.44 & 8.96 & 35.76 & \textbf{66.48} & \textbf{21.00}  & 40.48  & 36.19                   \\
		SOSNet+~\cite{SOSNet}                  & 52.76   & 31.48 & 32.68 & 8.76 & 35.44 & 65.20 & 20.76  & 39.64  & 35.84                   \\
		Exp-TLoss~\cite{ExpLoss}                & 52.68   & \textbf{31.68} & 32.76 & 9.24 & 35.44 & 63.36 & 20.92  & 39.28  & 35.67                   \\
		DSM+~\cite{DSM}                     & \textbf{52.96}   & \textbf{31.68} & \textbf{33.92} & 9.24 & 34.88 & 65.28 & 20.00  & 39.36  & 35.91                   \\
		TCDesc-HN+               & 52.80   & 31.52 & 33.60 & \textbf{9.72} & \textbf{35.52} & 65.40 & 20.47  & \textbf{40.76}  & \textbf{36.23}                   \\
		TCDesc-DSM+              & 52.88   & 31.64 & 33.12 & 9.16 & 35.36 & 64.52 & 20.36  & 40.64  & 35.96                   \\ \hline
	\end{tabular}
	\label{table_oxford}
\end{table*}

\subsection{Extensive Experiments}
\label{Extensive_Experiments}

\subsubsection{UBC PhotoTourism benchmark}
As illustrated above, UBC PhotoTourism~\cite{UBC} consists of \textit{Liberty}, \textit{Notredame} and \textit{Yosemite} three subsets, while we train on one subset and test on other two.
In the former section, we only train on subset \textit{Liberty}.
And in this section, we conduct our experiments on all tasks.

We test our method on the basis of HardNet~\cite{HardNet} and DSM~\cite{DSM}, which are named as \textbf{TCDesc-HN} and \textbf{TCDesc-DSM} respectively.
Specifically, we modify the distance of positive sample in their triplet losses as the adaptive weighting of Euclidean distance and topology distance of matching descriptors.
We set $k$ to be $16$, $\gamma$ to be $1.0$ during training CNNs.
We compare our method with SIFT~\cite{SIFT}, DeepDesc~\cite{DeepDesc}, L2-Net~\cite{L2Net}, HardNet~\cite{HardNet}, DOAP~\cite{DOAP}, ESE~\cite{ESE}, SOSNet~\cite{SOSNet}, Exp-TLoss~\cite{ExpLoss} and DSM~\cite{DSM}.
We present the performance of descriptors learned by various algorithms in Table.~\ref{UBCbenchmark}.

As can be seen, our novel topology measure improves performance of both descriptors learned by HardNet and DSM.
Specifically, mean FPR95 of HardNet declines from 1.51 to 1.22 and that of DSM declines from 1.18 to 1.08 after introducing our topology measure.
Furthermore, our method reduces the FPR95 of HardNet and DSM on every testing task.
Otherwise, as presented in Table.~\ref{UBCbenchmark}, our TCDesc on the basis of DSM leads the state-of-the-art result with the lowest FPR95 1.08.
The experimental results on UBC PhotoTourism benchmark validate the generalization of our method: We can improve performances of several descriptors learned by former triplet loss.

\subsubsection{HPatches benchmark}
In this section, we test the performance of our \textbf{TCDesc-HN} and \textbf{TCDesc-DSM} on HPatches benchmark~\cite{HPatches}.
We use model trained on subsets \textit{Liberty} of UBC PhotoTourism benchmark to generate descriptors from image patches of HPatches.
We compare our topology consistent descriptors \textbf{TCDesc-HN} and \textbf{TCDesc-DSM} with SIFT~\cite{SIFT}, HardNet~\cite{HardNet}, DOAP~\cite{DOAP}, SOSNet~\cite{SOSNet}, Exp-TLoss~\cite{ExpLoss} and DSM~\cite{DSM}, where our descriptors TCDesc-HN and TCDesc-DSM are trained on the basis of HardNet~\cite{HardNet} and DSM~\cite{DSM} respectively.

As can be seen in Fig.~\ref{HPatches}, there only exists a small margin among mAP of various learning-based descriptors in three tasks.
In task \textit{Patch Verification}, our TCDesc-DSM performs a little worse than DSM, and TCDesc-HN performs better than HardNet.
In task \textit{Image Matching}, our TCDesc-DSM and TCDesc-HN lead the state-of-the-art results and perform much better than DSM and TCDesc-HN, which proves the effectiveness of our topology consistent descriptors in image matching.
In task \textit{Patch Retrieval}, our TCDesc-DSM and TCDesc-HN both outperform than DSM and TCDesc-HN, and the TCDesc-DSM achieves the highest mAP (70.50) in this task.

\subsubsection{Wide baseline stereo}
Wide baseline stereo matching~\cite{wide_baseline} aims to find correspondences of two images in wide baseline setups, i.e., cameras with distant focal centers. So it is more challenging than normal image matching.
To verify generalization of our TCDesc and prove its advantages in extreme conditions, we conduct our experiments on W1BS benchmark~\cite{WxBS}.
W1BS dataset consists of 40 image pairs divided into 5 parts by the nuisance factors: \textit{Appearance}, \textit{Geometry}, \textit{Illumination}, \textit{Sensor} and \textit{Map to photo}.

W1BS datase uses multi detectors MSER~\cite{MSER}, Hessian-Affine~\cite{Hessian-Affine} and FOCI~\cite{FOCI} to detect affine-covariant regions and normalize the regions to size $41 \times 41$.
The average recall on ground truth correspondences of image pairs are employed to evaluate the performance of descriptors.

We compare our \textbf{TCDesc-HN} and \textbf{TCDesc-DSM} with SIFT~\cite{SIFT}, HardNet~\cite{HardNet}, SOSNet~\cite{SOSNet}, Exp-TLoss~\cite{ExpLoss} and DSM~\cite{DSM}.
Like the former experiment, we use the model trained on subsets \textit{Liberty} of UBC PhotoTourism benchmark to generate descriptors.
The experimental results are presented in Fig.~\ref{wxbs} where the larger mAUC indicates the better performance.
The mean mAUC of our TCDesc-HN is 8.78\%, which denotes the state-of-the-art performance.
And the mean mAUC of our TCDesc-DSM is 8.30\%, which is larger than that of DSM 8.12\%.
Conclusion could be drawn that our method can also improve performance of descriptors in extreme condition.

\subsubsection{Image Matching on Oxford Dataset}
Oxford dataset~\cite{Oxford} presents the real image matching scenarios, which takes in  many nuisance factors including blur, viewpoint change, light change and compression.
Oxford dataset~\cite{Oxford} only consists of 64 images but it is widely used to evaluate the robustness of image matching.
In this paper, we detect feature points by Harris-Affine~\cite{Hessian-Affine} detector, and we extract no more than 500 feature points for each image.
So we define the matching score as the right matches divided by $500$, where the lager matching score denotes the better performance.
We compare our \textbf{TCDesc-HN} and \textbf{TCDesc-DSM} with HardNet~\cite{HardNet}, SOSNet~\cite{SOSNet}, Exp-TLoss~\cite{ExpLoss} and DSM~\cite{DSM}.
We did not compare our method with SIFT~\cite{SIFT} because SIFT~\cite{SIFT} can not extract a fixed number of feature points as Harris-Affine~\cite{Hessian-Affine}.

We present our experimental results in Table.~\ref{table_oxford}.
As can be seen, our TCDesc-HN leads the state-of-the-art result ($36.23\%$ matching score) and our TCDesc-DSM+ ($35.91\%$ matching score) outperform than DSM+ ($35.96\%$ matching score).

\section{Conclusions}
\label{conclusion}
We observe the former triplet loss fails to maintain the similar topology between two descriptor sets since it takes the point-to-point Euclidean distance among descriptors as the only measure.
Inspired by the idea of neighborhood consistency of feature points in image matching, we try to learn neighborhood topology consistent descriptors by introducing a novel topology measure.

We first propose the linear combination weight to depict the topological relationship between center descriptor and its $k$NN descriptors, which is taken as the local topology weights.
We then propose the global mapping function which maps the local topology weights to the global topology vector.
Topology distance between two matching descriptors is defined as the $l1$ distance between their topology vector.
Last we propose the adaptive weighting strategy to jointly minimize topology distance and Euclidean distance of matching descriptors.
Experimental results on several benchmarks validate the generalization of our method since our method can improve performance of several algorithms using triplet loss.

However, our method is not appropriate for learning binary descriptors because the binary descriptor can not be linearly fitted by its $k$NN descriptors with float fitting weights.
We note that the idea of our method, local or neighborhood consistency can be extended to many other fields like cross-modal retrieval and etc.

\section*{Acknowledgment}
This study was supported by the National Natural Science Foundation of China (Grant No.61672183), by the Shenzhen Research Council (Grant No.JCYJ2017041310455226946,
JCYJ201708 15113552036), and in part by funding from Shenzhen Institute of Artificial Intelligence and Robotics for Society (Grant No.2019-INT021).

\bibliographystyle{IEEEtran}
\bibliography{IEEEabrv,mybibfile}


\begin{IEEEbiography}[{\includegraphics[width=1in,height=1.25in,clip,keepaspectratio]{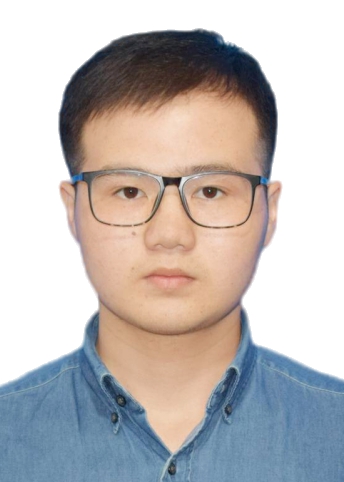}}]{Honghu Pan} received a master degree from Chongqing University in 2019. He is currently pursuing a Ph.D degree with the Department of Computer Science, Harbin Institute of Technology (Shenzhen). His research interests include computer vision, image matching and metric learning.
\end{IEEEbiography}

\begin{IEEEbiography}[{\includegraphics[width=1in,height=1.25in,clip,keepaspectratio]{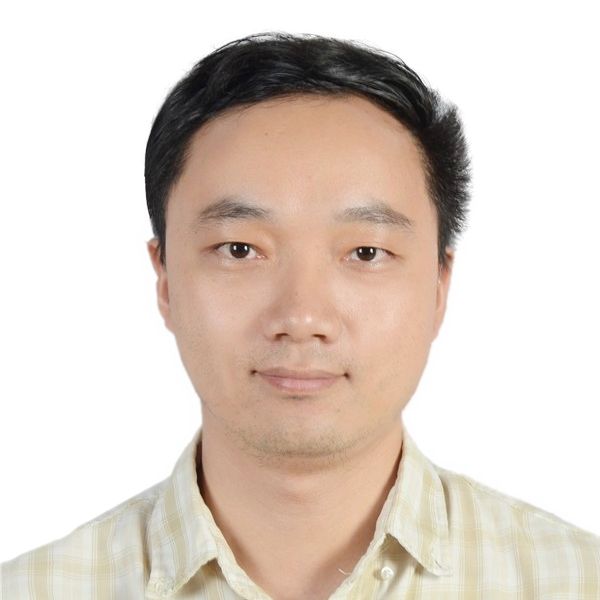}}]{Fanyang Meng} received the Ph.D. degree from the School of EE\&CS, Shenzhen University, China, in 2016, under the supervision of Prof. X. Li. He currently serves as a assistant research fellow in Peng Cheng Laboratory. His research interests include video coding, computer vision, human action recognition, and abnormal detection using RGB, depth, and skeleton data..
\end{IEEEbiography}

\begin{IEEEbiography}[{\includegraphics[width=1in,height=1.25in,clip,keepaspectratio]{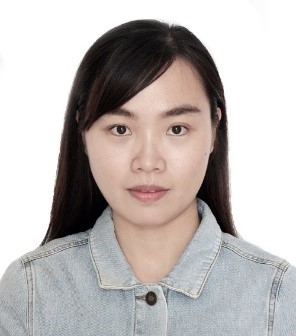}}]{Nana Fan} received the MS degree from the Harbin Institute of Technology Shenzhen Graduate School. She is currently pursuing the PhD degree with the Department of Computer Science, Harbin Institute of Technology (Shenzhen). Her research interests include visual tracking, deep learning, and machine learning.
\end{IEEEbiography}

\begin{IEEEbiography}[{\includegraphics[width=1in,height=1.25in,clip,keepaspectratio]{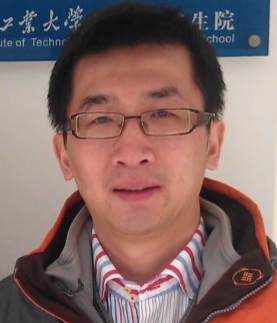}}]{Zhenyu He} received his Ph.D. degree from the Department of Computer Science, Hong Kong Baptist University, Hong Kong, in 2007. From 2007 to 2009, he worked as a postdoctoral researcher in the department of Computer Science and Engineering, Hong Kong University of Science and Technology. He is currently a full professor in the School of Computer Science and Technology, Harbin Institute of Technology, Shenzhen, China. His research interests include machine learning, computer vision, image processing and pattern recognition.
\end{IEEEbiography}

\vfill

\end{document}